\documentclass[journal]{IEEEtran}
\usepackage{booktabs}
\usepackage{multirow}
\usepackage{hyperref}
\usepackage{amsmath}
\usepackage{amssymb}
\usepackage{graphicx}
\usepackage{caption}
\usepackage{subcaption}
\usepackage{algorithm}
\usepackage{algorithmic}
\usepackage{color}
\usepackage[T1]{fontenc}
\usepackage{breakurl}

\newcommand{\f}{\mathbf{f}}
\newcommand{\g}{\mathbf{g}}
\newcommand{\h}{\mathbf{h}}
\newcommand{\n}{\mathbf{n}}
\newcommand{\M}{{\mathcal{M}}}

\title{A Nuclear-norm Model for Multi-Frame Super-Resolution Reconstruction from Video Clips}

\author{$^\ast$Rui~ Zhao ~and ~$^\dagger$Raymond ~H. ~Chan
\thanks{This work is partially supported
           by HKRGC GRF Grant No.
CUHK2130401, HKRGC CRF Grant No. CUHK2/CRF/11G, HKRGC CRF Grant No. C1007-15G,
HKRGC AoE Grant No.  AoE/M-05/12, CUHK
DAG No. 4053007, and CUHK FIS Grant No. 1902036.}
\thanks{$^\ast$R. Zhao is with Department of Mathematics, The Chinese University of Hong Kong, Shatin, NT, Hong Kong (Email: rzhao@math.cuhk.edu.hk).}
\thanks{$^\dagger$R. H. Chan is with Department of Mathematics, The Chinese University of Hong Kong, Shatin, NT, Hong Kong (Email: rchan@math.cuhk.edu.hk, Fax: (852) 2603-5154, Tel: (852) 3943-7970).}
}

\begin{document}

\maketitle

\begin{abstract}
We propose a variational approach to obtain super-resolution images from
multiple low-resolution frames extracted from video clips. First the
displacement between the low-resolution frames and the reference
frame are computed by an optical flow algorithm. Then a low-rank model
is used to construct the reference frame in high-resolution by incorporating the information of the low-resolution frames. The model
has two terms: a 2-norm data fidelity term and a nuclear-norm regularization term. Alternating direction method of multipliers  is used to solve the model.
Comparison of our methods with other models on synthetic and real video clips show that
our resulting images are more accurate with less artifacts. It also provides much finer and discernable details.
\end{abstract}

\section{Introduction} \label{section1}
\IEEEPARstart{S}uper-resolution (SR) image reconstruction from multiple low-resolution (LR) frames have many applications, such as in remote sensing, surveillance, and medical imaging. After the pioneering work of Tsai and Huang \cite{tsai1984multiframe}, SR image reconstruction
has become more and more popular in image
processing community, see for examples \cite{bose1998high,shankar2008sparsity,shen2004biorthogonal,farsiu2006multiframe,chan2007framelet,lu2007multi,duponchel2008super,takeda2009super}.
SR image reconstruction problems can be classified into two categories:
single-frame super-resolution problems (SFSR) and multi-frame super-resolution problems (MFSR). In this paper, we mainly focus on the multi-frame case, especially the MFSR problems from low-resolution video sequences. Below, we first review some existing work related to MFSR problems.

Bose and Boo \cite{bose1998high} considered the case where the multiple LR image frames were shifted with affine transformations. They modeled the original high-resolution (HR)
image as a stationary
Markov-Gaussian random field. Then they made use of the {maximum a posteriori} scheme to solve their model. However the affine transformation assumption may not be satisfied in practice, for example when there are complex motions or  illumination changes.
Another approach for SR image reconstruction is the one known as patch-based or learning-based. Bishop \textit{et al.} \cite{bishop2003super} used a set of learned image patches which capture the information between the middle and high spatial frequency bands. They assumed a priori distribution over such patches and made use of the previous enhanced frame to provide part of the training set. The disadvantage of this patch-based method is that it is usually time consuming and sensitive to the off-line training set.
Liu and Sun \cite{liu2014bayesian} applied Bayessian approach to estimate simultaneously the underlying motion, the blurring kernel, the noise level and the HR image. Within each iteration, they estimated the motion, the blurring kernel and the HR image alternatively by maximizing a posteriori respectively. Based on this work, Ma \textit{et al}. \cite{ma2015multi} tackled motion blur in their paper. An expectation maximization (EM) framework is applied to the Bayessian approach to guide the estimation of motion blur. These methods used optical flow to model the motion between different frames. However they are sensitive to the accuracy of flow estimation. The results may fail when the noise is heavy.

In \cite{chan2003wavelet}, Chan \textit{et al}. applied wavelet analysis to HR image reconstruction.
They decomposed the image from previous iteration into wavelet frequency domain and applied wavelet thresholding to denoise the resulting images.
Based on this model,  Chan \textit{et al}. \cite{chan2004tight} later developed an iterative MFSR approach by using tight-frame wavelet filters. However because of the number of framelets involved in analyzing the LR images, the algorithm can be extremely time consuming.

Optimization models are one of the most important image processing models. Following the classical ROF model \cite{rudin1992nonlinear}, Farsiu \textit{et al}. \cite{farsiu2004fast} proposed a total variation-$l_1$ model
where they used the $l_1$ norm for the super-resolution data fidelity term. However it is known that TV regularization enforces a piecewise solution. Therefore their method will produce some artifacts. Li, Dai and Shen \cite{li2010multiframe} used $l_1$ norm of the geometric tight-framelet coefficients as the regularizer and adaptively mimicking $l_1$ and $l_2$ norms as the data fidelity term. They also assumed affine motions between different frames. The results are
therefore not good when complex motions or illumination changes are involved.

Chen and Qi \cite{chen2013single} recently proposed a single-frame HR image reconstruction method via low rank regularization. Jin \textit{et al}. \cite{chen2013video} designed a patch based low rank matrix completion algorithm from the sparse representation of LR images.
The main idea of these two papers  is based on the assumption that each LR image is downsampled from a blurred and shifted HR image. However these work assumed that the original HR image, when considered as
a matrix, has a low rank property, which is not convincing in general.

In this paper, we show that the low rank property can in fact be constructed under MFSR framework. The idea is to consider each LR image as a downsampled instance of a
{\it different} blurred and shifted HR image. Then when all these
different HR images are properly aligned, they should give a low rank matrix; and therefore
we can use a low rank prior to obtain a better solution.
Many existing work assumes the shift between two consecutive LR frames are
small, see, e.g.,
 \cite{altunbasak2002superresolution,farsiu2004fast,wang2006improved,narayanan2007computationally,zibetti2007robust}.
 In this paper, we allow illumination changes and more complex motions other than affine
transformation. They
are handled by an optical flow model proposed in \cite{Gilliam2015Local}.
Once the motions are determined, we reconstruct the high-resolution image
by minimizing a functional which consists of two terms: the 2-norm data fidelity term
to suppress Gaussian noise and a nuclear-norm regularizer to
enforce the low-rank prior. Tests on seven synthetic and real video clips
show that our resulting images is more accurate with less artifacts.  It can also provide
much finer and discernable details.

The rest of the paper is organized as follows.
Section \ref{sec:lowresolutionmodel} gives a brief
review of a classical model on modeling LR images from HR images.
Our model will be based on this model.
Section \ref{sec:nuclearmodel} provides the details of
our low-rank model, including image registration by optical flow
and the solution of our optimization problem by alternating
direction method. Section \ref{sec:experiments} gives experimental results on
the test videos. Conclusions are given in Section \ref{sec:conclusion}.

To simplify our discussion, we now give the notation that we will be using for the rest of the paper. For any integer $m\in\mathbb{Z}$, $I_{m}$ is
the $m\times m$ identity matrix. For any integer $l \in \mathbb{Z}$ and positive integer $n\in\mathbb{Z}^+$, there exists a unique $0\leq\widetilde{l}<n$ such that $\widetilde{l}\equiv l \mod n$. Let $N_n(l)$ denote the $n\times n$ matrix
\begin{equation}\label{equ:IntegerMotionMat}{N}_n(l)=
\left[
\begin{array}{cc}0&I_{n-\widetilde{l}}\\I_{\widetilde{l}}&0\end{array}
\right].
\end{equation}
For a vector $\f\in\mathbb{R}^{n}$, $N_n(l)\f$ is the vector
with entries of $\f$ cyclic-shifted by $l$.

Define the downsampling matrix $D_i$ and the upsampling matrix $D^T_i$ as
\begin{equation}
\label{equ:down-up-matrices}
D_i(n)=I_n\otimes\mathbf{e}_i^T \text{ and } D^T_i(n)=I_n\otimes \mathbf{e}_i, ~i=0, 1,
\end{equation}
where $\mathbf{e}_0=[1,0]^T, \mathbf{e}_1=[0,1]^T$ and $\otimes$ is the Kronecker product. For $0\leq\epsilon\leq 1$, define $T_n(\epsilon)$ to be the $n\times n$ Toeplitz matrix
\begin{equation}\label{equ:toeplitzmatrix}
T_n(\epsilon)=\left[\begin{array}{ccccc}
1-\epsilon & \epsilon & \cdots &0 \\
0& 1-\epsilon & \ddots& \vdots\\
\vdots & \ddots & \ddots &\epsilon\\
\epsilon &\cdots&0 & 1-\epsilon
\end{array}
\right].
\end{equation}
This Toeplitz matrix performs the effect of linear interpolation shifted by $\epsilon$.

\section{Low resolution model with shifts}
\label{sec:lowresolutionmodel}

Consider a LR sensor array recording a video of an object.
Then it gives multiple LR images of the object. Unless
the object or the sensor array is completely motionless during the
recording, the LR images will contain multiple information of the object
at different shifted locations (either because of the motion of
the object or of the sensor array itself). Our problem is to
improve the resolution of one of the LR images (called the
reference image) by incorporating information from the
other LR images.

\begin{figure}[htp!]
	\centering
	\begin{subfigure}[b]{0.30\textwidth}
		\centering
		\includegraphics[width=\textwidth]{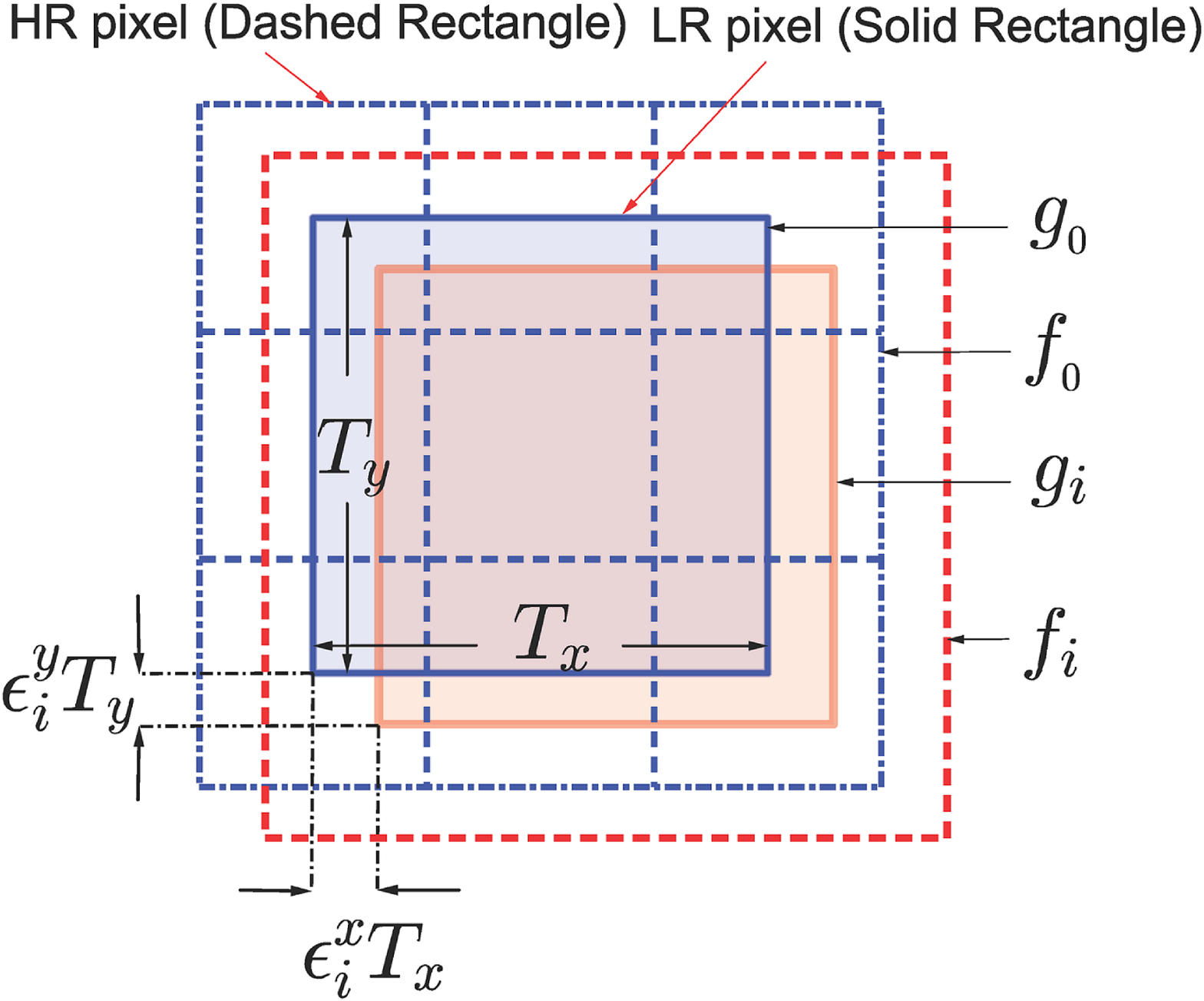}
		\caption{Displacements between LR images}
		\label{fig:LRHR}
	\end{subfigure}
	\begin{subfigure}[b]{0.26\textwidth}
		\centering
        \includegraphics[width=\textwidth]{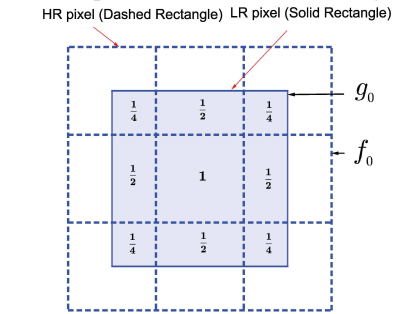}
		\caption{The averaging process }
		\label{fig:LRHR-2}
	\end{subfigure}
	\caption{LR Images with displacements}
\end{figure}

Let the sensor array consist of $m\times n $ sensing elements,
where the width and the height of each sensing element is $L_x$ and $L_y$
respectively.  Then, the sensor array will produce an $m\times n$ discrete image with $mn$ pixels where each of these LR pixels is of size $L_x\times L_y$. Let $r$ be the upsampling factor, i.e. we would like
to construct an image of resolution $rm \times rn$ of the same scene.
Then the size of the HR pixels will be $L_x/r\times L_y/r$.
Fig \ref{fig:LRHR} shows an example. The big rectangles with solid edges
are the LR pixels and the small rectangles with dashed edges are the HR pixels.

Let $\{g_i\in\mathbb{R}^{m\times n},  1\leq i \leq p\}$ be the sequence of LR images produced by the sensor array at different time points, where $p$ is the number of frames. For simplicity we let $g_{0}$ be the reference LR image which
can be chosen to be any one of the LR images $g_i$.
The displacement of $g_i$ from the reference image $g_0$ is
denoted by $(\epsilon^x_iL_x, \epsilon^y_iL_y)$, see the solid rectangle in Fig. \ref{fig:LRHR} labeled as $g_i$. For ease of notation, we
will represent the 2D images $g_i$, $0\le i \le p$, by vectors $\g_i\in\mathbb{R}^{mn}$
obtained by stacking the columns of $g_i$. We use $\mathbf{f}\in\mathbb{R}^{r^2mn}$ to denote
the HR reconstruction of $g_0$ that we are seeking.

We model the relationship between $\f$ and $\g_0$ by averaging, see
\cite{bose1998high,chan2007framelet}.  Fig. \ref{fig:LRHR-2} illustrates that
the intensity value of the LR pixel is the weighted average of the
intensity values of the HR pixels overlapping with it. The weight is precisely the area of overlapping. Thus the process from
$\f$ to each of the LR images $g_i$ can be modeled by \cite{chan2007framelet}:
\begin{equation}
\label{equ:HR2LR}
\mathbf{g}_i=DKA_i \mathbf{f}+\mathbf{n}_i, ~i=1, 2, \cdots, p,
\end{equation}
where
$D=D_0(n)\otimes D_0(m)\in\mathbb{R}^{mn\times r^2mn}
$ is the downsampling matrix defined by (\ref{equ:down-up-matrices}); $K\in\mathbb{R}^{r^2mn\times r^2mn}$ is the average
operator mentioned above; $A_i\in\mathbb{R}^{r^2mn\times r^2mn}$ is the warping matrix which measures the displacement between $g_i$ and $g_0$; and  $\mathbf{n}_i$ is the additive unknown noise. In this paper, we assume for simplicity the noise are Gaussian.
Other noise models can be handled by choosing suitable data fidelity terms.

The warping matrix $A_i$, $1\le i \le p$, is to align the
LR pixels in $\g_i$ at exactly the middle of the corresponding HR pixels in $\f$,
exactly like the $\g_0$ is w.r.t $\f_0$ in Fig. \ref{fig:LRHR-2}. Once this alignment is done, the average operator $K$, which is just a blurring operator,
can be written out easily. In fact, the 2D kernel (i.e. the point spread function)
of $K$ is given by $vv^T$, where $v=[1/2, 1, \ldots, 1, 1/2]^T$ with $(r-1)$ ones in the middle, see \cite{bose1998high}.
The $A_i$ are more difficult to obtain. In the most ideal case where the motions are only translation of less than one HR pixel length and width, $A_i$ can be modeled by $A_i = T_n(\epsilon_i^x)\otimes T_m(\epsilon_i^y)$, where $T_n(\epsilon_i^x), T_m(\epsilon_i^y)$ are Toeplitz matrices given by (\ref{equ:toeplitzmatrix}) with $(\epsilon^x_iL_x,\epsilon^y_iL_y)$ being the horizontal and vertical displacements of $g_i$, see Fig. \ref{fig:LRHR} and \cite{chan2007framelet}. In reality, the changes between different LR frames are much more complicated. It can involve illumination changes and other complex non-planar motions. We will discuss the formation of $A_i$ in more details in
Subsections \ref{s3.1} and \ref{sub:imageregistration}.

\section{Nuclear model}
\label{sec:nuclearmodel}

Given (\ref{equ:HR2LR}), a way to obtain $\f$ is to minimize
the noise $\n_i$ by least-squares. However because $D$ is
singular, the problem is ill-posed. Regularization is necessary to make use of some priori information to choose the correct solution. A typical regularizer for solving this problem is \textit{Total Variation} (TV)  \cite{rudin1992nonlinear}. The TV model is well known for edge preserving and  can give a reasonable solution for MFSR problems. However it assumes that the HR image is piecewise constant. This will produce some artifacts.

Instead we will develop a low-rank model for the problem.
The main motivation is as follows. We consider each LR image $\g_i$ as a downsampled
version of an HR image $\f_i$. If all these HR images
$\f_i$ are properly aligned with the HR image $\f$, then they all
should be the same exactly (as they are representing the same
scene $\f$). In particular, if $A_i$ is the
alignment matrix that aligns $\f_i$ with $\f$, then the matrix
$[A_1 \f_1,A_2 \f_2, \ldots, A_p\f_p]$ should be a low rank matrix
(ideally a rank 1 matrix). Thus the rank of the matrix can be used as a prior.

In Subsection \ref{s3.1}, we introduce our low-rank model in the case where the
LR images are perturbed only by translations.
Then in Subsection \ref{s3.2}, we explain how to solve the model
by the alternating direction method. In Subsection \ref{sub:imageregistration},
we discuss how to modify the model when there are more complex
motions or changes between the LR frames.

\subsection{Decomposition of the warping matrices} \label{s3.1}

In order to introduce our model without too cumbersome notations,
we assume first here that the displacements of the LR images
from the reference frame are translations only.
Let $s_i^xL_x$ and $s_i^yL_y$ be the horizontal and vertical
displacements of $g_i$ from $g_0$.
(How to obtain $s_i^x$ and $s_i^y$ will be discussed
in Subsection  \ref{sub:imageregistration}.)
Since the width and height of one
HR pixel are $L_x/r$ and $L_y/r$ respectively, the
displacements are equivalent to $rs_i^x$ HR pixel length and $rs_i^y$ HR pixel width.
We decompose $rs_i^x$ and $rs_i^y$ into the integral parts and fractional parts:
\begin{equation}
\label{equ:translation}
rs_i^x=l_i^x+\epsilon_i^x,  ~rs_i^y=l_i^y+\epsilon_i^y,
\end{equation}
where $l_i^x, l_i^y$ are integers and $0\leq \epsilon_i^x, \epsilon_i^x< 1$. Then the warping matrix can be decomposed as:
\begin{equation}\label{equ:warpingmatrices}
A_i=C_iB_i,
\end{equation}
where $B_i = N_n(l_i^x)\otimes N_m(l_i^y)$ is given by  (\ref{equ:IntegerMotionMat}) and $C_i = T_n(\epsilon_i^x)\otimes T_m(\epsilon_i^y)$ is given by (\ref{equ:toeplitzmatrix}) \cite{chan2003wavelet}. Thus by
letting $\f_i=B_i\f$, $1 \le i \le p$, (\ref{equ:HR2LR}) can be rewritten as
\begin{equation}
\label{equ:HR2LR1}
\mathbf{g}_i=DKC_i \mathbf{f}_i+\mathbf{n}_i, ~i=1, 2, \cdots, p.
\end{equation}

\begin{figure}[htp!]
\centering
        \centering
        \includegraphics[width=0.4\textwidth]{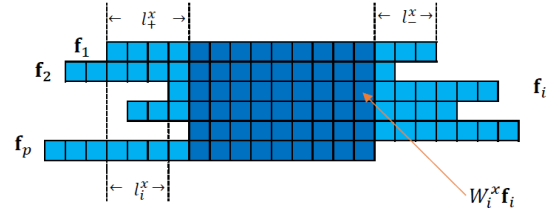}
    \caption{$1$-D signals with integer displacements}
    \label{fig:fig3}
\end{figure}

As mentioned in the motivation above, all these
$\f_i$, which are equal to $B_i\f$, are integral shift from $\f$. Hence if they
are aligned correctly by an alignment matrix
$W_i$, then the overlapping entries
should be the same. Fig.  \ref{fig:fig3} is the 1D
illustration of this idea. The $W_i^x$ is the matrix
that aligns $\f_i$ with $\f$ (in the $x$-direction)
and the dark squares are the overlapping pixels
and they should all be the same as the corresponding pixels in $\f$.

Mathematically, $W_i$ is constructed as follows.
Given the decomposition of $rs_i^x$ and $rs_i^y$ in (\ref{equ:translation}), let $l_+^x=\max_i\{0, l_i^x\}$, $l_+^y=\max_i\{0, l_i^y\}$,  and $l_-^x=\max_i\{0, -l_i^x\}$,  $l_-^y=\max_i\{0, -l_i^y\}$. Then
\begin{equation}
\label{equ:wmatrix}
W_i=W_i^x
\otimes
W_i^y.
\end{equation}
where $$
W_i^x=\left[
\begin{array}{ccc}
0_{l_+^x-l_i^x}&&\\
&I_{rn-l_+^x-l_-^x}&\\
&&0_{l_-^x+l_i^x}
\end{array}
\right],$$$$W_i^y=
\left[
\begin{array}{ccc}
0_{l_+^y-l_i^y}&&\\
&I_{rm-l_+^y-l_-^y}&\\
&&0_{l_-^y+l_i^y}
\end{array}
\right].$$
Note that $W_i$ nullifies the entries outside the overlapping part
(i.e. outside the dark squares in Fig.  \ref{fig:fig3}).

Ideally, the matrix $[W_1\f_1,  W_2\f_2,  \cdots , W_p\f_p]$ should be
a rank-one matrix as every column should be a replicate of $\f$ in the
overlapping region. In practice, it can be of low rank due to various
reasons such as errors in measurements and noise in the given video. Since nuclear norm is the
convexification of low rank prior, see \cite{candes2011robust}, this leads to our convex model
\begin{equation}
\label{equ:nuclear}
\min_{\f_1, \cdots, \f_p}\lambda\|W_1\f_1,  W_2\f_2 , \cdots , W_p\f_p\|_*+\frac{1}{2}\sum_{i=1}^p\|\g_i-DKC_i\f_i\|_2^2,
\end{equation}
where $\|\cdot\|_*$ is  the matrix nuclear norm and $\lambda$ is the
regularization parameter.  We call our model (\ref{equ:nuclear}) the \textit{nuclear model}.
We remark that here we use the 2-norm data fidelity term because we assume
the noise are Gaussian. It can be changed to another norm according to the noise type.

\subsection {Algorithm for solving the nuclear model} \label{s3.2}

We  use \textit{alternating direction method of multipliers} (ADMM) \cite{boyd2011distributed} to solve the nuclear model. We replace  $\{W_i\f_i\}_{i=1}^p$ in the model by variables $\{\h_i\}_{i=1}^p$.  Let  $H=[\h_1, \h_2, \cdots, \h_p]$,  $F=[\f_1, \f_2, \cdots, \f_p]$,  and $WF=[W_1\f_1, W_2\f_2, \cdots, W_p\f_p]$. The \textit{Augmented Lagrangian} of model (\ref{equ:nuclear}) is
\begin{eqnarray*}
\mathcal{L}_{\lambda\rho}(H, F, \Lambda)&=&\lambda\|H\|_*+\frac{1}{2}\sum_{i=1}^p\|\g_i-DKC_i\f_i\|^2_2\\
&+&\sum_{i=1}^p\langle \Lambda_i, \h_i-W_i\f_i\rangle+\frac{1}{2\rho}\|H-WF\|^2_{{\cal{F}}},
\end{eqnarray*}
where $\Lambda=[\Lambda_1, \Lambda_2, \cdots, \Lambda_p]$ is the Lagrangian multiplier,
$\| \cdot \|_{{\cal{F}}}$ is the Frobenius norm, and $\rho$ is an algorithm parameter.

To solve the nuclear model,  it is equivalent to minimize $\mathcal{L}_{\lambda\rho}$, and we use ADMM \cite{boyd2011distributed} to minimize it. The idea of the scheme is to minimize $H$ and $F$ alternatively by fixing the other,  i.e.,  given the initial value $ F^0, \Lambda^0$,  let
$	H^{k+1}=\arg\min_H\mathcal{L}_{\lambda\rho}(H, F^k, \Lambda^k),$
$	F^{k+1}=\arg\min_F\mathcal{L}_{\lambda\rho}(H^{k+1}, F, \Lambda^k),$
 where $k$ is the iteration number.
These two problems are  convex  problems.  The \textit{singular value threshold} (SVT) gives the solution of the $H$-subproblem.  The $F$-subproblem is reduced to
solving $p$ linear systems.  For a matrix $X$,  the SVT of $X$ is defined to be $SVT_\rho(X)=U\Sigma_\rho^+V^T$
where $X=U\Sigma V^T$ is the singular value decomposition (SVD) of $X$ and $\Sigma_\rho^+=\max\{\Sigma-\rho, 0\}$. We summarize the algorithm in Algorithm \ref{alg:nuclear2} below. It is well-known that the algorithm is
convergent if $\rho > 0$ \cite{boyd2011distributed}.

\begin{algorithm}
 	\caption{$\f \leftarrow (\{g_i,W_i,C_i\},K,  \lambda, \rho, \Lambda^0, F^0)$}
 	\label{alg:nuclear2}
 	\begin{algorithmic}
 		\FOR { $k=1, 2, 3, \cdots$}
 		\STATE  $H^{k+1}=SVT_{\lambda\rho}(WF^k-\Lambda^k)$;
 		\FOR {$i=1$ to $p$}
        \STATE $M_i=(DKC_i)^TDKC_i+\frac{1}{\rho}W_i^TW_i$;
 		\STATE $\f_i^{k+1}=\left(M_i\right)^{-1}\left((DKC_i)^T\g_i+W_i^T\Lambda_i^k+\frac{1}{\rho}W_i^T\h_i^{k+1}\right)$;
 		\ENDFOR
 		\STATE  $\Lambda^{k+1}=\Lambda^{k}+\frac{1}{\rho}(H^{k+1}-WF^{k+1})$;
 		\ENDFOR
 		\STATE Output: $\f$ as the the average of the columns of $F^k$.
 	\end{algorithmic}
 \end{algorithm}

In Algorithm \ref{alg:nuclear2},  the $SVT$ operator involves the SVD of a  matrix $WF^k-\Lambda^k$. The number of its columns  is $p$, the
number of LR frames, which is relatively small.  Therefore the SVT step is not time consuming. For the second subproblem, we need to solve $p$ linear systems. The coefficient matrices contain some structures which help accelerating the calculation. The matrices $D^TD$ and $W_i^TW_i$  are diagonal matrices while $K$ and $C_i$ can be diagonalized by either FFT
or DCT depending on the boundary conditions we choose, see \cite{ng1999fast}. In our tests, we always use periodic boundary conditions.

\subsection{Image registration and parameter selection}
\label{sub:imageregistration}

In Algorithm \ref{alg:nuclear2},  we assume that there are   only translations  between different LR frames. However  there can be other complex motions and/or illumination changes in practice.  We handle these by using the \textit{Local All-Pass} (LAP) optical flow algorithm proposed in \cite{Gilliam2015Local}.
Given a set of all-pass filters $\{\phi_j\}_{j=0}^N$ and $\phi:=\phi_0+\sum_{j=1}^{N-1}c_j \phi_j$, the optical flow $\M_i$ of $g_i$
is obtained by solving the following problem:
$$
\min_{\{c_1,\cdots,c_{N-1}\}}\sum_{l,k\in R}|\phi{{(k,l)}} g_i(x-k,y-l)-\phi(-k,-l)g_{0}(x-k,y-l)|^2,
$$
where $R$ is a window centered at $(x,y)$. In our experiments, we followed the settings in the paper \cite{Gilliam2015Local}, and let  $N=6, R=16$ and
\begin{eqnarray*}
&\phi_0(k,l)=e^{-\frac{k^2+l^2}{2\sigma^2}}, \quad & \phi_1(k,l)=k\phi_0(k,l),\\
 &\phi_2(k,l)=l\phi_0(k,l), \quad& \phi_3(k,l)=(k^2+l^2-2\sigma^2)\phi_0(k,l),\\
 &\phi_4(k,l)=kl\phi_0(k,l),\quad &\phi_5(k,l)=(k^2-l^2)\phi_0(k,l),
\end{eqnarray*}
where $\sigma=\frac{R+2}{4}$ and $\phi$ is supported in $[-R,R]\times[-R,R]$.
The coefficients $c_n$ can be obtained by solving a linear system.  The optical flow $\M_i$ at $(x,y)$ is then given by
$$\M_i(x,y)=\left(\frac{2\sum_{k,l}k\phi(k,l)}{\sum_{k,l}\phi(k,l)}, \frac{2\sum_{k,l}l\phi(k,l)}{\sum_{k,l}\phi(k,l)}\right),$$
which can be used to transform $g_i$ back to the grid of $g_0$.
In order to increase the speed by avoiding interpolation, here we consider only
the integer part of the flow.  Hence we get the {\textit{restored LR images }}
\begin{equation}
	\label{equ:flowtransform}
	\tilde{g}_i(x,y)=g_i([\M_i](x,y)), ~i=1,2,\cdots,p, \quad \forall (x,y) \in \Omega
\end{equation}
where $[\M_i]$ is the integer part of the flow $\M_i$ and $\Omega$ is the image domain.

The optical flow handles complex motions and illumination changes and will restore the positions of pixels in $g_i$ w.r.t $g_0$. To enhance the accuracy of the image registration, we further estimate if there
are any translation that are unaccounted for after the optical flow. In particular, we assume that $\tilde{g}_i$ may be displaced from $g_0$ by a simple translation
\begin{equation}
\label{equ:affinetransform}\mathcal{T}(x, y)
=\left[\begin{array}{c}x\\y\end{array}\right]
-\left[\begin{array}{c}s_i^x\\ s_i^y\end{array}\right].
\end{equation}
To estimate the displacement vector  $[s_i^x, s_i^y]^T$,
we use the Marquardt-Levenberg algorithm proposed in  \cite{thevenaz1998pyramid}. It aims to minimize the squared error
\begin{equation}\label{equ:motionerrors}
E(\tilde{g}_i, g_0)=\sum_{(x, y)\in\Omega}[\tilde{g}_i(\mathcal{T}(x, y))-g_0(x, y)]^2.
\end{equation}
The detail implementation of this algorithm can be found
in \cite[Algorithm 3]{chan2007framelet}.
After obtaining $[s^x_i, s^y_i]$, then by (\ref{equ:warpingmatrices}) and (\ref{equ:wmatrix}),  we can construct the matrices $C_i$ and $W_i$ for our nuclear model (\ref{equ:nuclear}).

Before giving out the whole algorithm,  there remains the problem about parameters selection. There are two parameters to be determined: $\lambda$ the regularization parameter and $\rho$ the algorithm (ADMM) parameter. We need to tune these two  parameters in practice such that the two subproblems can be solved effectively and accurately. Theoretically, $\rho$ will not affect the minimizer of the model but only the convergence of the algorithm \cite{boyd2011distributed}. However in order to get an effective algorithm,  it should not be set very small.
For $\lambda$, we use the following empirical formula to approximate it in each iteration \cite{li2010multiframe},
\begin{equation}\label{equ:lambda}\lambda \approx \frac{1/2\sum_{i=1}^p\|\widetilde{\g}_i-DKC_i\f_i^k\|^2}{\|W_1\f_1^k,  W_2\f_2^k , \cdots , W_p\f_p^k\|_*},
\end{equation}
where $\f_i^k$ is the estimation of $\f_i$ in the $k$-th iteration. The formula may not give the best $\lambda$
but can largely narrow its scope. We then use trial and error to get the best parameter.
We give out the full algorithm for our model below.

\begin{algorithm}
	\caption{$\f \leftarrow (\{g_i\},i_0,K,\Lambda^0,F^0,\lambda, \rho)$  }
	\label{alg:nuclear3}
	\begin{algorithmic}
		\FOR{ $i=0, 1, 2, \cdots p$ }
		\STATE Compute $\widetilde{g}_i(x, y)$ from (\ref{equ:flowtransform});
		\STATE Compute $s^x_i$ and $s^y_i$ in (\ref{equ:affinetransform})
by using the Marquardt-Levenberg algorithm in \cite[Algorithm 3]{chan2007framelet}
		\STATE Compute the warping matrices $C_i$ and $W_i$,  according to (\ref{equ:warpingmatrices}) and (\ref{equ:wmatrix});
		\ENDFOR
		\STATE Apply Algorithm \ref{alg:nuclear2} to compute the HR images $\f \leftarrow (\{\widetilde{g}_i,W_i,C_i\},K,  \lambda, \rho, \Lambda^0, F^0)$;
		\STATE Output $\f$.
	\end{algorithmic}
\end{algorithm}

\section{Numerical experiments}
\label{sec:experiments}

In this section, we illustrate the effectiveness of our algorithm by
comparing it with 3 different variational methods on
7 synthetic videos and real videos. Chan \textit{et al}. \cite{chan2003wavelet} applied wavelet analysis to MFSR problem and then developed an iterative approach by using tight-frame wavelet filters \cite{chan2007framelet}. We refer their model as \textit{Tight Frame} model (TF). Li, Dai and Shen \cite{li2010multiframe} proposed the \textit{Sparse Directional Regularization} model (SDR) where they used $l_1$ norm of the geometric tight-framelet coefficients as the regularizer and the adaptively-mimicking $l_1$ and $l_2$ norms as the data fidelity term. Ma \textit{et al}. \cite{ma2015multi} introduced an expectation-maximization (EM) framework to the Bayessian approach of Liu and Sun \cite{liu2014bayesian}. They also tackled motion blur in their paper. We refer it as the MAP model.
We will compare our Algorithm \ref{alg:nuclear3} (the nuclear model)  with  these three methods. The sizes of the videos we used are listed in Table \ref{tab:basicinfoofvideos}. The CPU timing of all methods are also listed there. Except for one case (Eia with $r=2$)
our model is the fastest, see the boldfaced numbers there.

{There is one parameter  for the TF model---a thresholding parameter $\eta$ which controls the registration quality of the restored LR images $\tilde{g}_i$ (see (\ref{equ:flowtransform})). If the PSNR value between $\tilde{g}_i$ and the reference image $g_0$ are smaller than $\eta$, it will discard $\tilde{g}_i$ in the reconstruction. We apply \textit{trial and error} method to choose the best $\eta$. For the SDR method, we use the default setting in the paper \cite{li2010multiframe}. Hence the parameters are selected automatically by the algorithm. The TF model, the SDR model and the nuclear model are applied to  $\tilde{g}_i$, i.e. we use the same optical flow algorithm \cite{Gilliam2015Local} for these three models. For the MAP model, it utilized an optical flow algorithm from Liu \cite{liu2009beyond}. Following the paper, the optical flow parameter $\alpha$  is very small. We also apply \textit{trail and error} method to tune it.

All the videos used in the tests as well as the results are available at:
\noindent\url{http://www.math.cuhk.edu.hk/\textasciitilde rchan/paper/super-resolution/experiments.html}


\begin{table*}
	\caption{Size of each data set and CPU time for all models.}
	\label{tab:basicinfoofvideos}
	\begin{center}
		\begin{tabular}{ l | ccc |c | cccc}
			&\multicolumn{3}{c|}{Size of data}&Factor	 &\multicolumn{4}{|c}{CPU time (in seconds) }\\
			\hline				
			& Height&Width&Frame&$r$&TF&MAP	&SDR	&Nuclear\\
			\hline
			Boat	&240&240&17&2&3470& 252		&	119&{\bf 78}	\\ \hline
			Boat	&120&120&17&4&18518& 212		&	124&{\bf 67	}\\ \hline
			Bridge&240&240&17	&2&3954&	261	&		127&{\bf87}	\\ \hline
			Bridge&120&120&17	&4&22641&	209	&		125&{\bf 63}	\\ \hline
			Text	&57&49&21&2&1583&	23	&7.6	&{\bf 6.1}\\ \hline
			Text	&57&49&21&4&10601&	42	&19	&{\bf 10}\\ \hline
			Disk	&57&49&19&2&1243&	21	&7.4		&{\bf5.4}\\ \hline
			Disk	&57&49&19&4&13469&	40	&19		&{\bf 10}\\ \hline
			Alpaca&96&128&21&2	&2146&59	&21	&{\bf16	}\\ \hline
			Alpaca&96&128&21&4	&25233&188	&105	&{\bf 57	}\\ \hline
      Eia & 90&90& 16 &2&1854 & 33& {\bf 8.2} & 8.8   \\ \hline
			Eia & 90&90& 16 &4& 36034 & 61& 56 & {\bf 26  } \\ \hline
			Books &288&352&21&2	&9265 &	614	&830	&{\bf 606}	
		\end{tabular}
	\end{center}
\end{table*}

\subsection{Synthetic videos}
We start from a given HR image $\f^*$, see e.g. the boat image in Fig. \ref{fig:boat512-HR}.
We translate and rotate $\f^*$ with known parameters and also change their illuminations by different scales. Then we downsample these frames with the given
factor $r=2$ or $r=4$ to get our LR frames $\{\g_i\}_{i=1}^p$.
We take $p=17$, and Gaussian noise of ratio 5\% is added to each LR frame.

After we reconstruct the HR image $\f$ by a method, we compare it with the true solution $\f^*$ using two popular error measurements. The first one is \textit{peak signal-to-noise ratio} (PSNR) and the second one is \textit{structural similarity } (SSIM) \cite{wang2004ssim}. For two signals $\mathbf{x}=(x_1, x_2, \cdots, x_n)^T$ and $\mathbf{y}=(y_1, y_2, \cdots, y_n)^T$, they are defined by
$$\text{PSNR}(\mathbf{x}, \mathbf{y})=10\log_{10}\left(\frac{d^2}{\|\mathbf{x}-\mathbf{y}\|^2/n}\right), $$$$\text{SSIM}(\mathbf{x}, \mathbf{y})=\frac{(2\mu_x\mu_y+c_1)(2\sigma_{xy}+c_2)}{(\mu_x^2+\mu_y^2+c_1)(\sigma_x^2+\sigma_y^2+c_2)},$$
where $d$ is the dynamic range of $\mathbf{x}, \mathbf{y}$ and $\mu_x, \mu_y$ are
the mean values of $\mathbf{x}$ and $\mathbf{y}$; $\sigma_x, \sigma_y$ are
the variances of $\mathbf{x}$ and $\mathbf{y}$; $\sigma_{xy}$ is the covariance of $\mathbf{x}$ and $\mathbf{y}$; $c_i, i=1, 2$ are constants related to $d$, which are typically set to be $c_1=(0.01d)^2, c_2=(0.03d)^2$. Because of the motions, we do not have enough information to reconstruct $\f$ near the boundary; hence
this part of $\f$ will not be accurate.
Thus we restrict the comparison within the overlapping area of all LR images.

Table \ref{tab:boatbridgepsnrssim} gives the PSNR values and SSIM values
of the reconstructed HR images $\f$ from the boat
and the bridge videos. The results show that our model gives much more
accurate $\f$ for both upsampling factor $r=2$ and $4$, see
the boldfaced values there. The improvement is significant when comparing
to the other three models, e.g. at least 1.6dB in PSNR for the boat video when $r=2$.
From Table \ref{tab:basicinfoofvideos}, we also see that our method is the fastest.
To compare the images visually, we give the results and their zoom-ins for the boat
video in Figs. \ref{fig:boat512nScl2results}--\ref{fig:subboat512nScl4results}.
The results for the bridge video are similar and therefore omitted.
Fig. \ref{fig:boat512nScl2results} shows the boat reconstructions for  $r=2$.
We notice that the TF model loses many fine details, e.g., the ropes of the mast.
The MAP model produces some distortion on the edges and is sensitive to the noise;
and the SDR model contains some artifacts along the edges. One can see the difference more clearly from the zoom-in
images in Fig. \ref{fig:subboat512nScl2results}. We also give the zoom-in results for $r=4$ in Fig. \ref{fig:subboat512nScl4results}. We can see that the nuclear model produces more details and less artifacts than the other three models.

\begin{table*}	
	\centering
	\caption{PSNR and SSIM values for the ``Boat'' and ``Bridge'' videos.}
	\label{tab:boatbridgepsnrssim}
		\begin{tabular}{ l|l | cccc |cccc}
            \multicolumn{2}{c|}{~}&\multicolumn{4}{|l|}{Upsampling factor $r=2$}&\multicolumn{4}{|c}{Upsampling factor $r=4$}\\ \hline							 &&TF&MAP&SDR&Nuclear &TF&MAP&SDR&Nuclear\\\hline
			\multirow{2}{*}{Boat}&PSNR &18.7&25.3&28.2&{\bf 29.8}&20.7&23.6&27.0	&{\bf 27.5}	\\ \cline{2-10}
			&SSIM&0.69&0.70&0.80&{\bf 0.83}&0.69&0.67&0.72&{\bf 0.77}\\ \hline
            \multirow{2}{*}{Bridge}&PSNR&20.7&23.6&27.0&{\bf27.5}&20.1&22.4&24.1&{\bf 25.0}\\	 \cline{2-10}
			&SSIM&0.69&0.67&0.72&{\bf 0.77}&0.53&0.57&0.65&{\bf 0.72}
		\end{tabular}
\end{table*}

\begin{figure*}[htp!]
  \centering
  \begin{subfigure}[t]{0.28\textwidth}
    \centering
    \includegraphics[width=\textwidth]{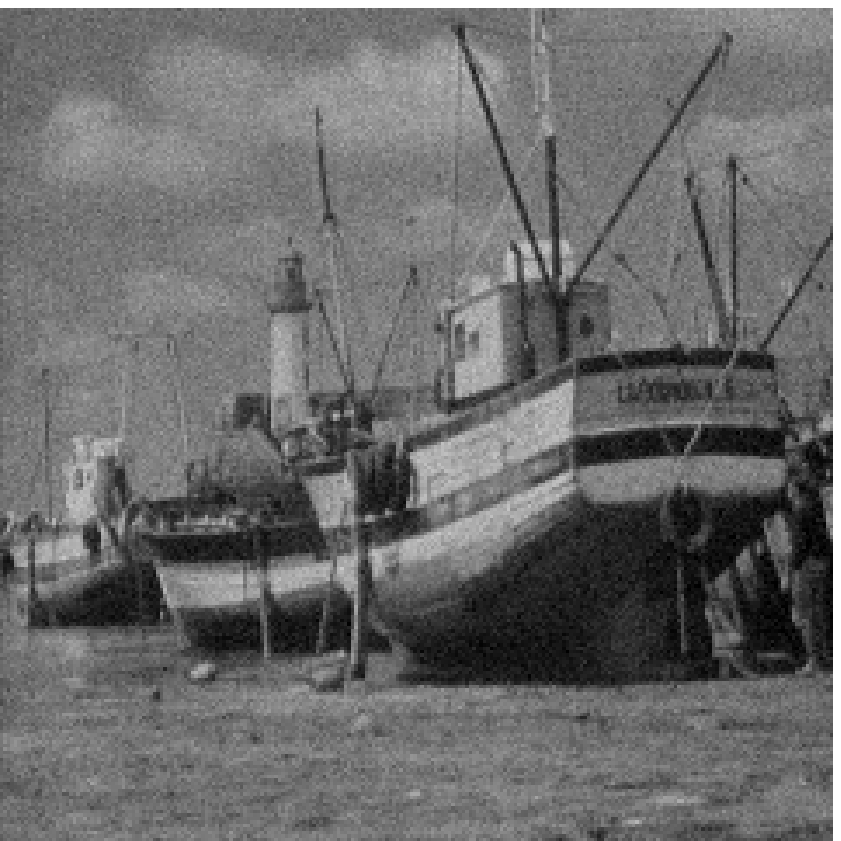}
    \caption{Reference LR image}
    \label{fig:boat512-LR}
  \end{subfigure}
  \begin{subfigure}[t]{0.28\textwidth}
    \centering
    \includegraphics[width=\textwidth]{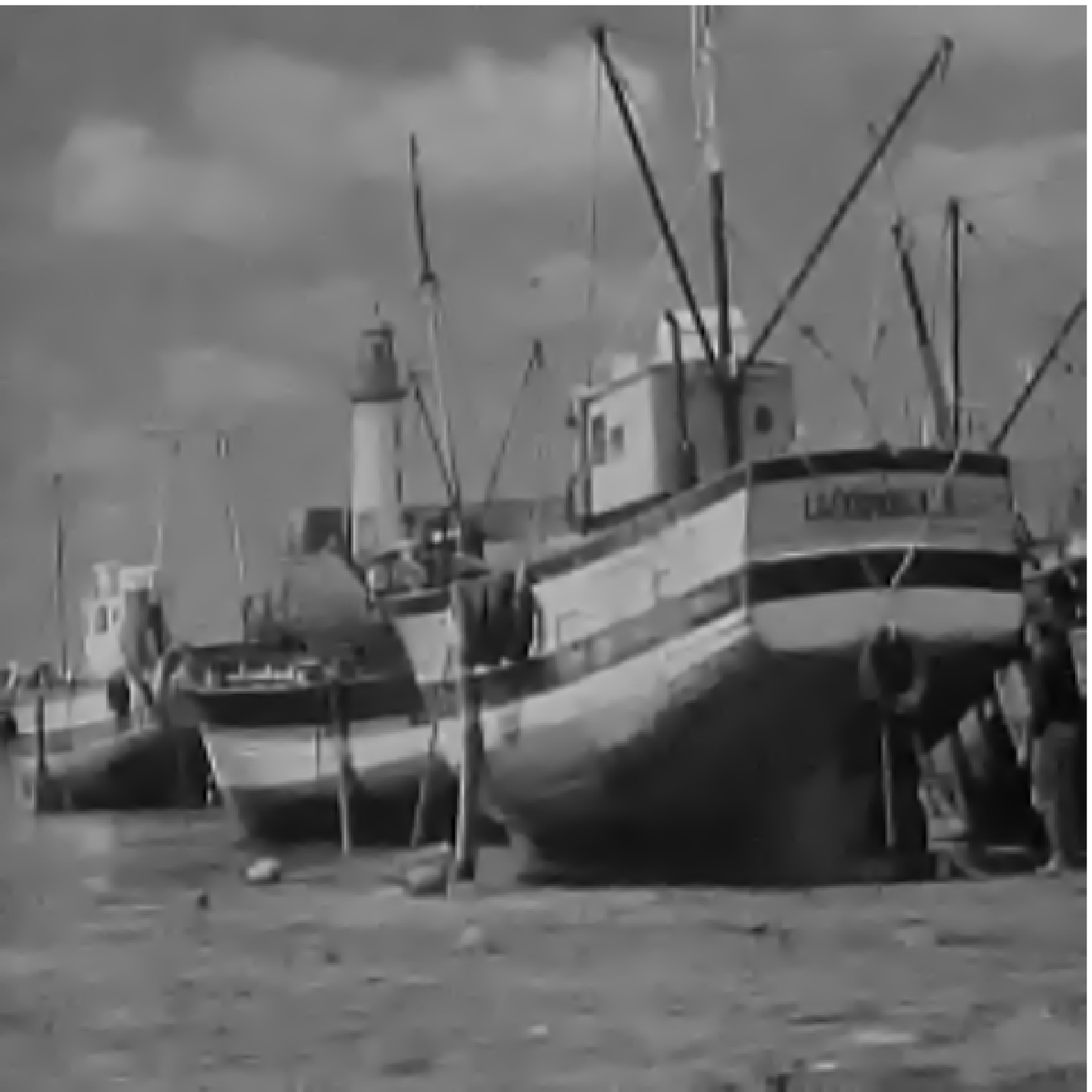}
    \caption{TF}
    \label{fig:boat512-TF}
  \end{subfigure}
	\begin{subfigure}[t]{0.28\textwidth}
		\centering
		\includegraphics[width=\textwidth]{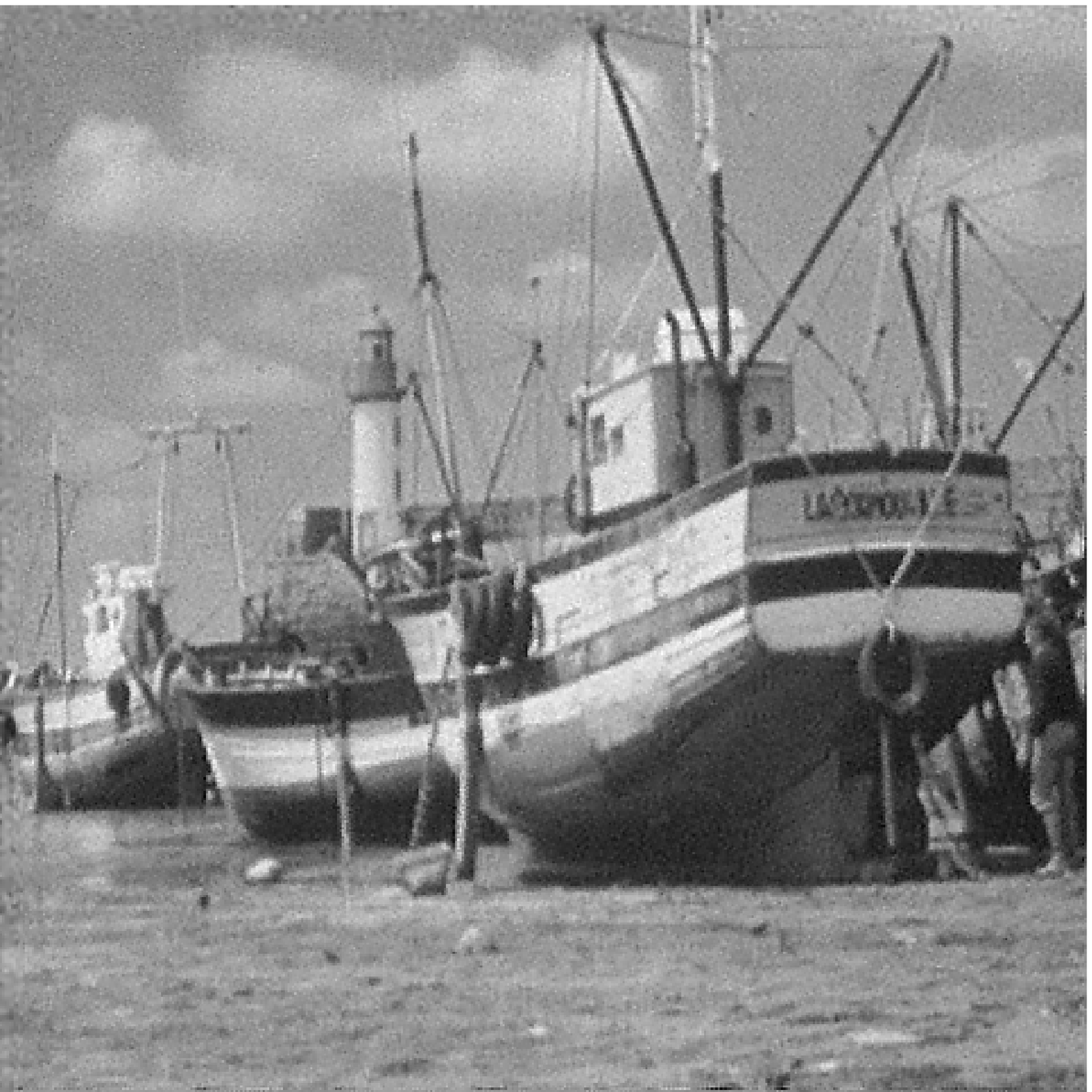}
		\caption{MAP}
		\label{fig:boat512-MZ}
	\end{subfigure}
  \begin{subfigure}[t]{0.28\textwidth}
    \centering
    \includegraphics[width=\textwidth]{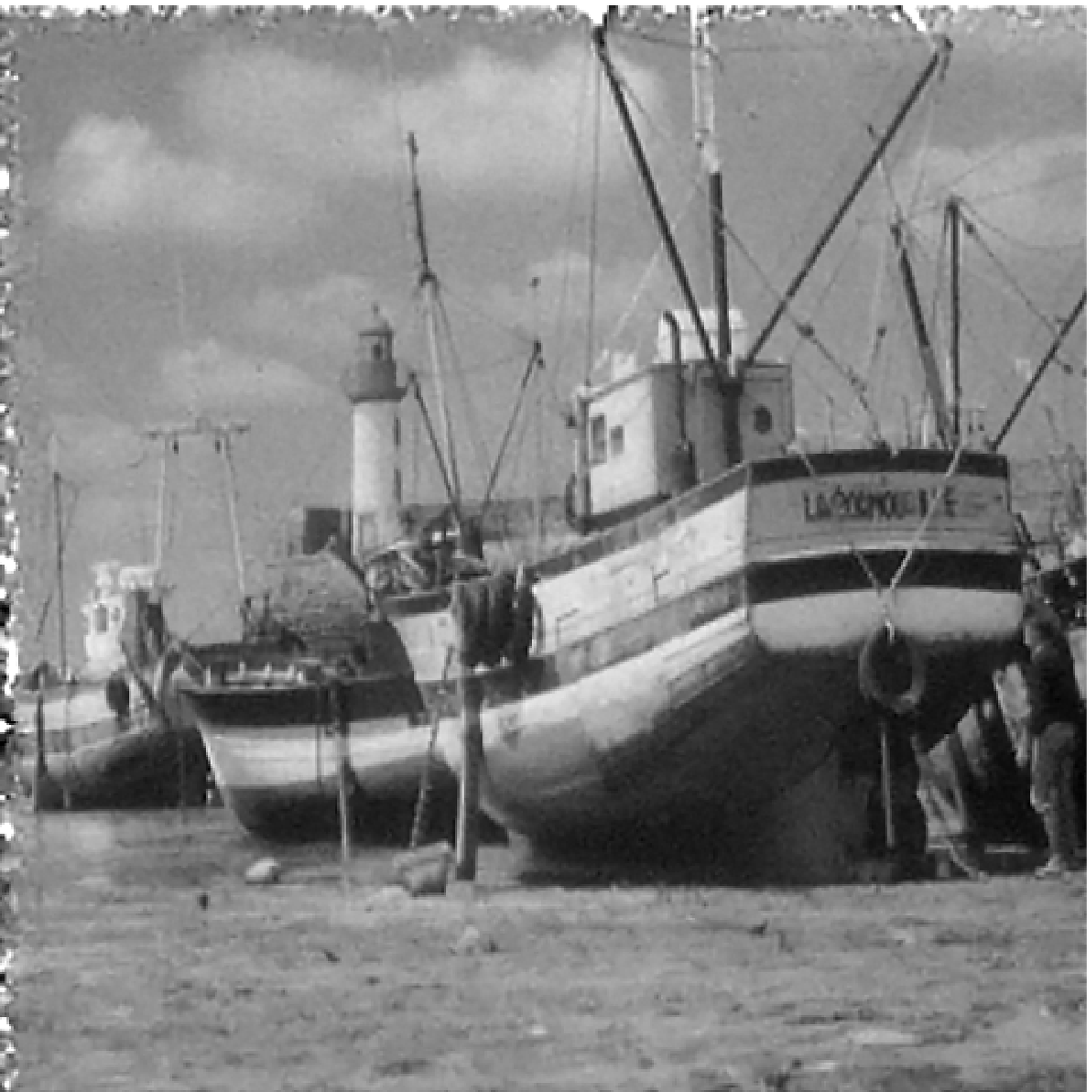}
    \caption{ SDR}
    \label{fig:boat512-SDR}
  \end{subfigure}
  \begin{subfigure}[t]{0.28\textwidth}
    \centering
    \includegraphics[width=\textwidth]{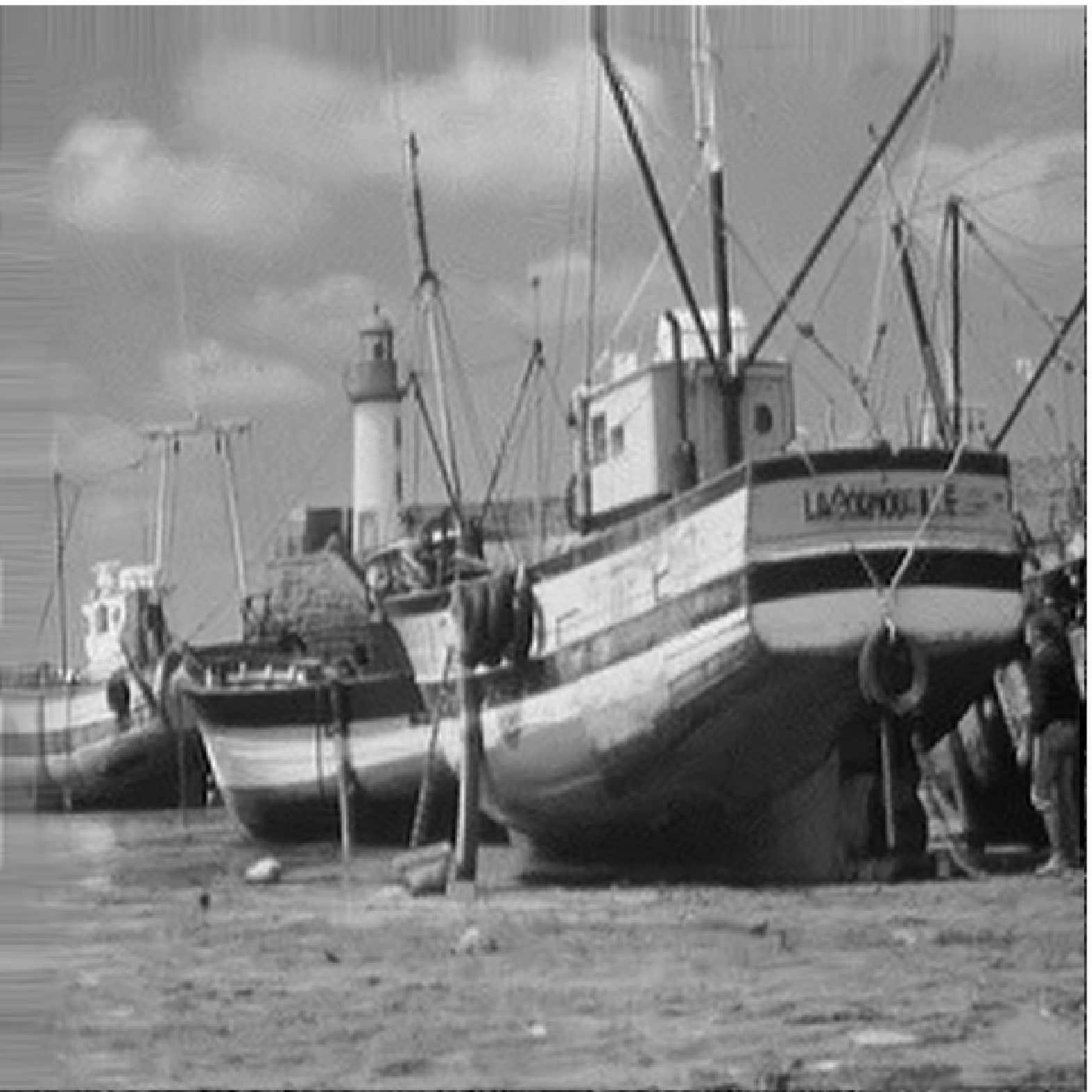}
    \caption{Nuclear}
    \label{fig:boat512-Nuclear5}
  \end{subfigure}
  \begin{subfigure}[t]{0.28\textwidth}
    \centering
    \includegraphics[width=\textwidth]{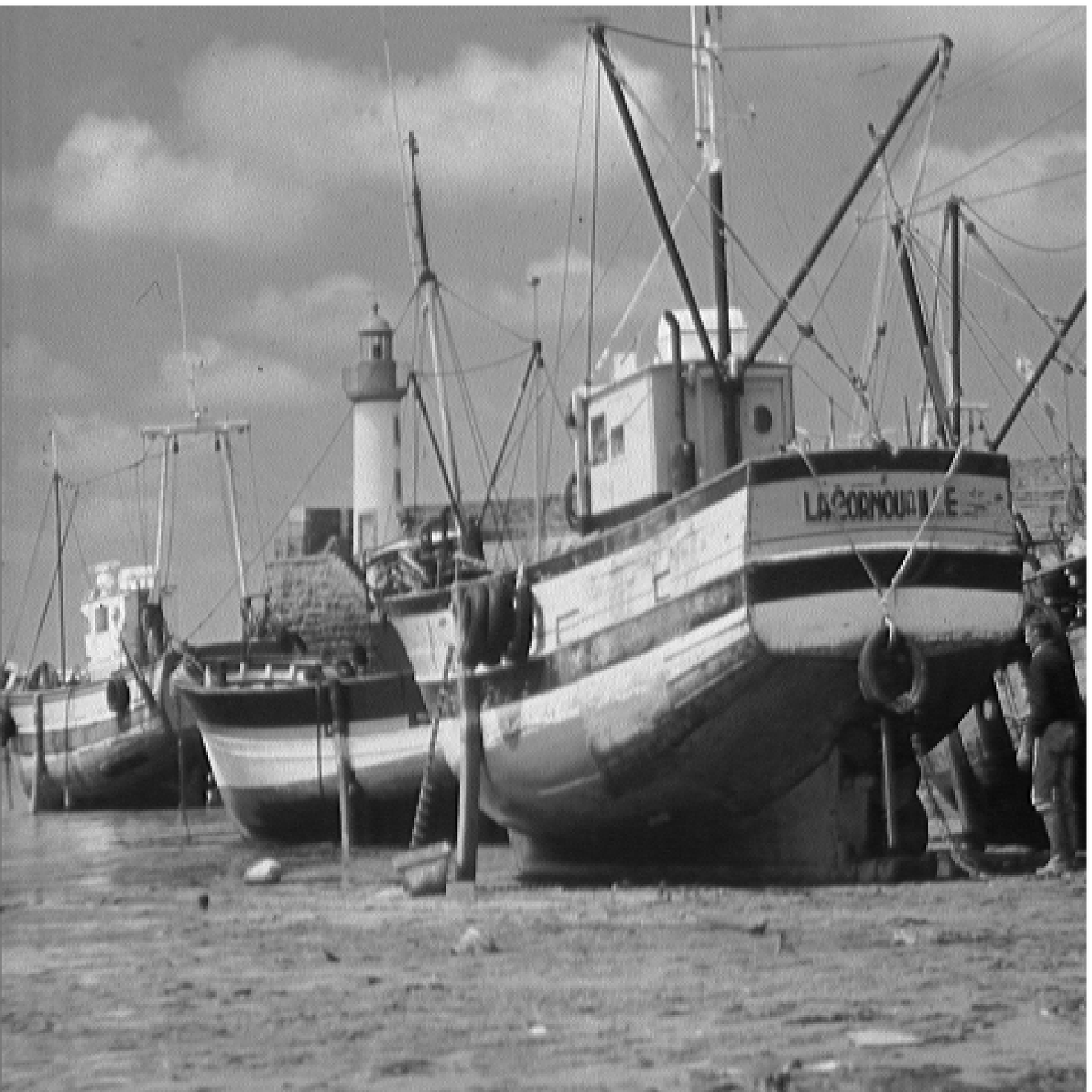}
    \caption{True HR image}
    \label{fig:boat512-HR}
  \end{subfigure}
   \caption{Comparison of different algorithms on ``Boat'' image
   with upsampling factor $r=2$.
  	(a) The reference LR image.
  	(b) Result of the TF model \cite{chan2007framelet}.
  	(c) Result of the MAP model \cite{ma2015multi}.
  	(d) Result of the SDR model \cite{li2010multiframe}.
  	(e) Result of our nuclear model ($\lambda=1, \rho=400$).
  	(f) True HR image.}
  \label{fig:boat512nScl2results}
\end{figure*}

\begin{figure*}[htp!]
	\centering
	\begin{subfigure}[t]{0.28\textwidth}
		\centering
		\includegraphics[width=\textwidth]{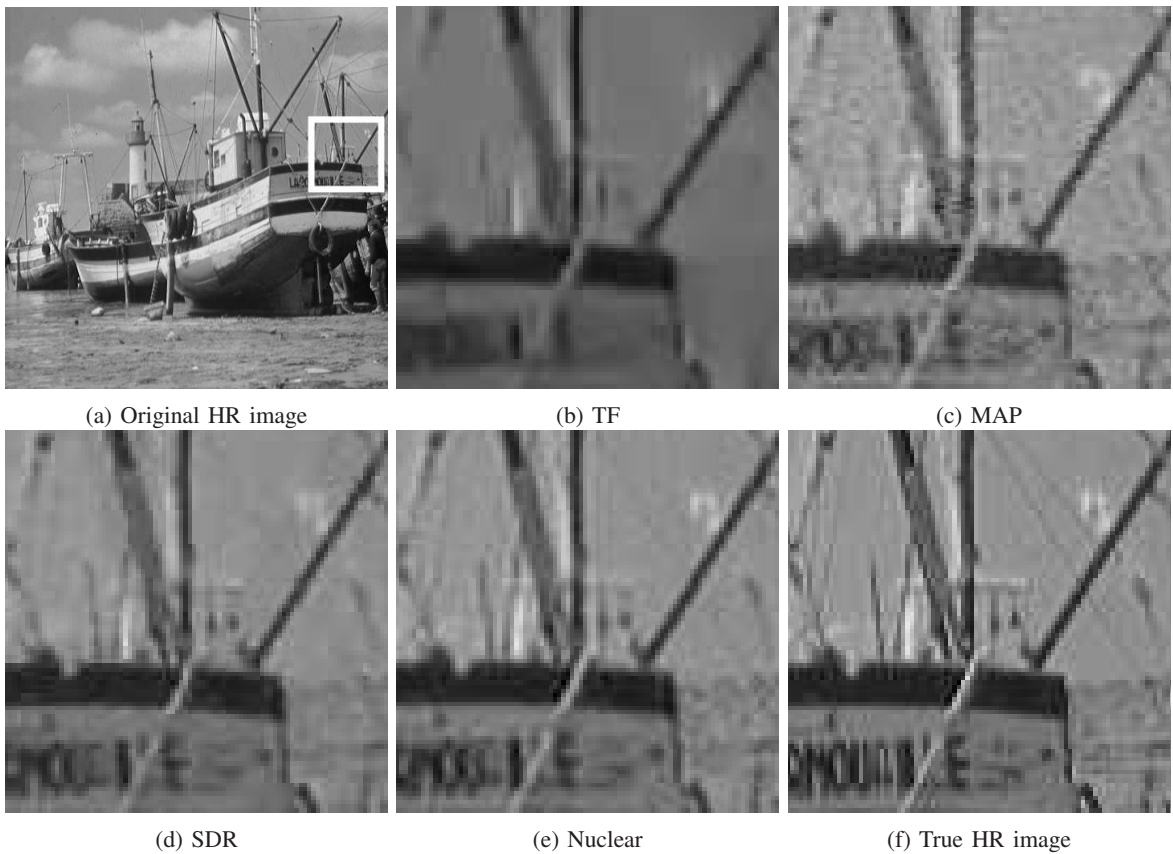}
		\caption{Original HR image}
		\label{fig:boat512-HR-Rec}
	\end{subfigure}
	\begin{subfigure}[t]{0.28\textwidth}
		\centering
		\includegraphics[width=\textwidth,trim={382pt 250pt 10pt 142pt} ,clip]{boat512-HR-TF.eps}
		\caption{ TF}
		\label{fig:subboat512-TF}
	\end{subfigure}
	\begin{subfigure}[t]{0.28\textwidth}
	\centering
	\includegraphics[width=\textwidth,trim={382pt 250pt 10pt 142pt} ,clip]{boat512-HR-MZ.eps}
	\caption{ MAP}
	\label{fig:subboat512-MZ}
\end{subfigure}
	\begin{subfigure}[t]{0.28\textwidth}
		\centering
		\includegraphics[width=\textwidth,trim={382pt 250pt 10pt 142pt},clip]{boat512-HR-SDR.eps}
		\caption{SDR}
		\label{fig:subboat512-SDR}
	\end{subfigure}
	\begin{subfigure}[t]{0.28\textwidth}
		\centering
		\includegraphics[width=\textwidth,trim={382pt 250pt 10pt 142pt},clip]{boat512-HR-Nuclear5.eps}
		\caption{Nuclear}
		\label{fig:subboat512-Nuclear5}
	\end{subfigure}
	\begin{subfigure}[t]{0.28\textwidth}
		\centering
		\includegraphics[width=\textwidth,trim={382pt 250pt 10pt 142pt},clip]{boat512-HR.eps}
		\caption{True HR image}
		\label{fig:subboat512-HR}
	\end{subfigure}
	\caption{Zoomed-in comparison of different algorithms on ``Boat'' image for $r=2$.
		(a) The zoom-in part in the HR image.
		(b) Result of the TF model \cite{chan2007framelet}.
		(c) Result of the MAP model \cite{ma2015multi}.
		(d) Result of the SDR model \cite{li2010multiframe}.
		(e) Result of our nuclear model ($\lambda=1, \rho=400$).
		(f) Zoomed-in original HR image.}
	\label{fig:subboat512nScl2results}
\end{figure*}

\begin{figure*}[htp!]
	\centering
	\begin{subfigure}[t]{0.28\textwidth}
		\centering
		\includegraphics[width=\textwidth]{boat512-LR.eps}
		\caption{Reference LR image}
		\label{fig:boat512nScl4-HR-Rec}
	\end{subfigure}
	\begin{subfigure}[t]{0.28\textwidth}
		\centering
		\includegraphics[width=\textwidth,trim={382pt 250pt 10pt 142pt} ,clip]{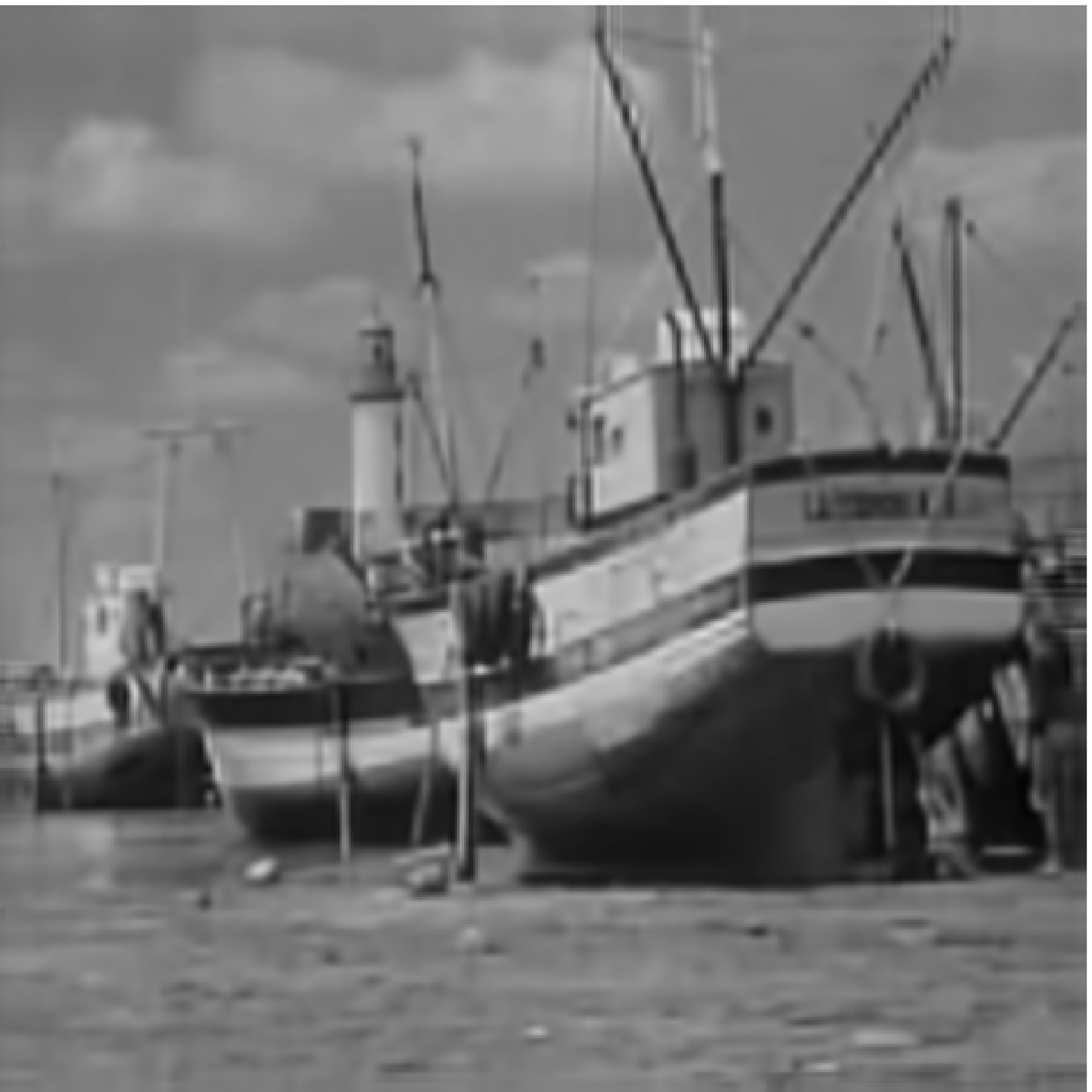}
		\caption{ TF}
		\label{fig:subboat512nScl4-TF}
	\end{subfigure}
	\begin{subfigure}[t]{0.28\textwidth}
		\centering
		\includegraphics[width=\textwidth,trim={382pt 250pt 10pt 142pt} ,clip]{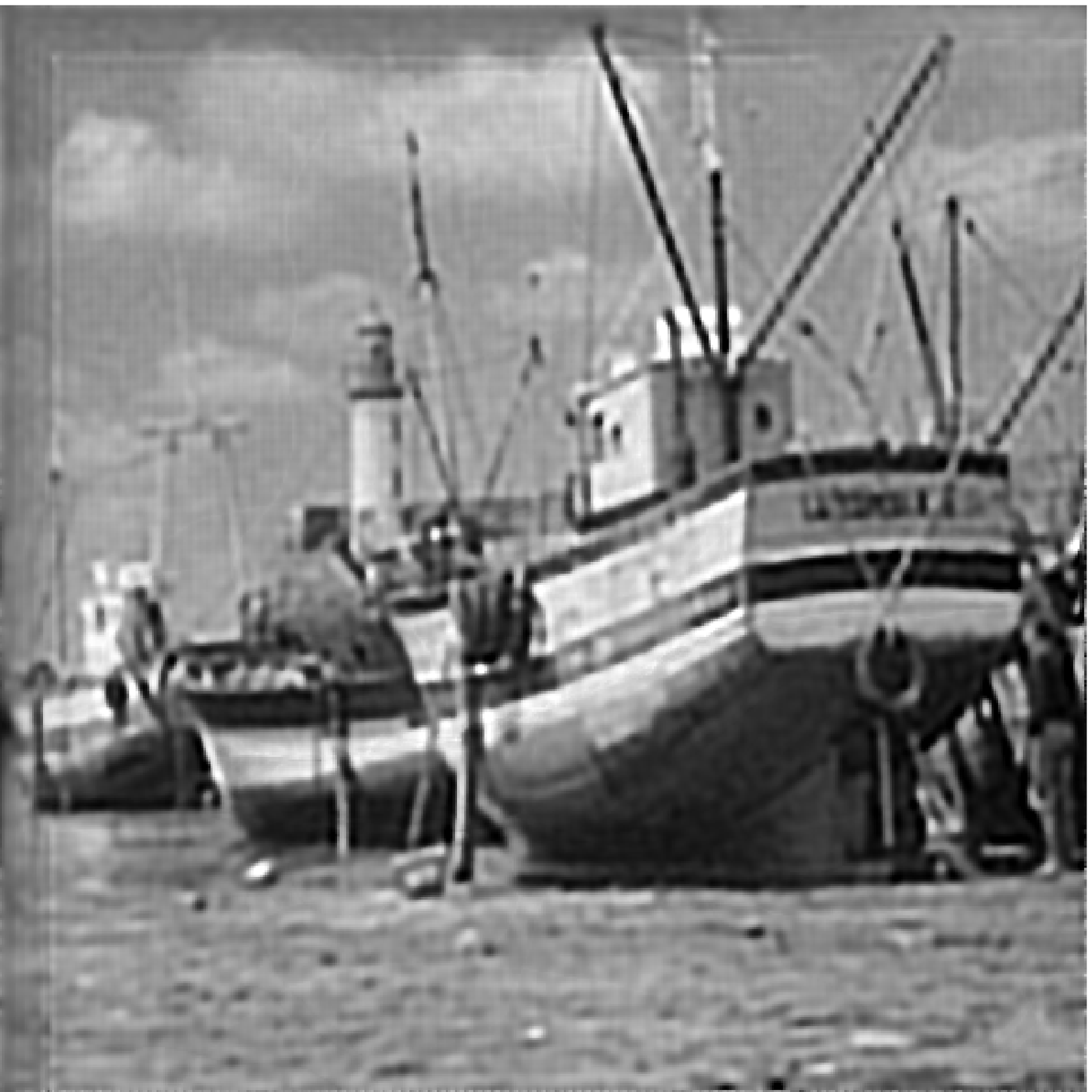}
		\caption{ MAP}
		\label{fig:subboat512nScl4-MZ}
	\end{subfigure}
	\begin{subfigure}[t]{0.28\textwidth}
		\centering
		\includegraphics[width=\textwidth,trim={382pt 250pt 10pt 142pt},clip]{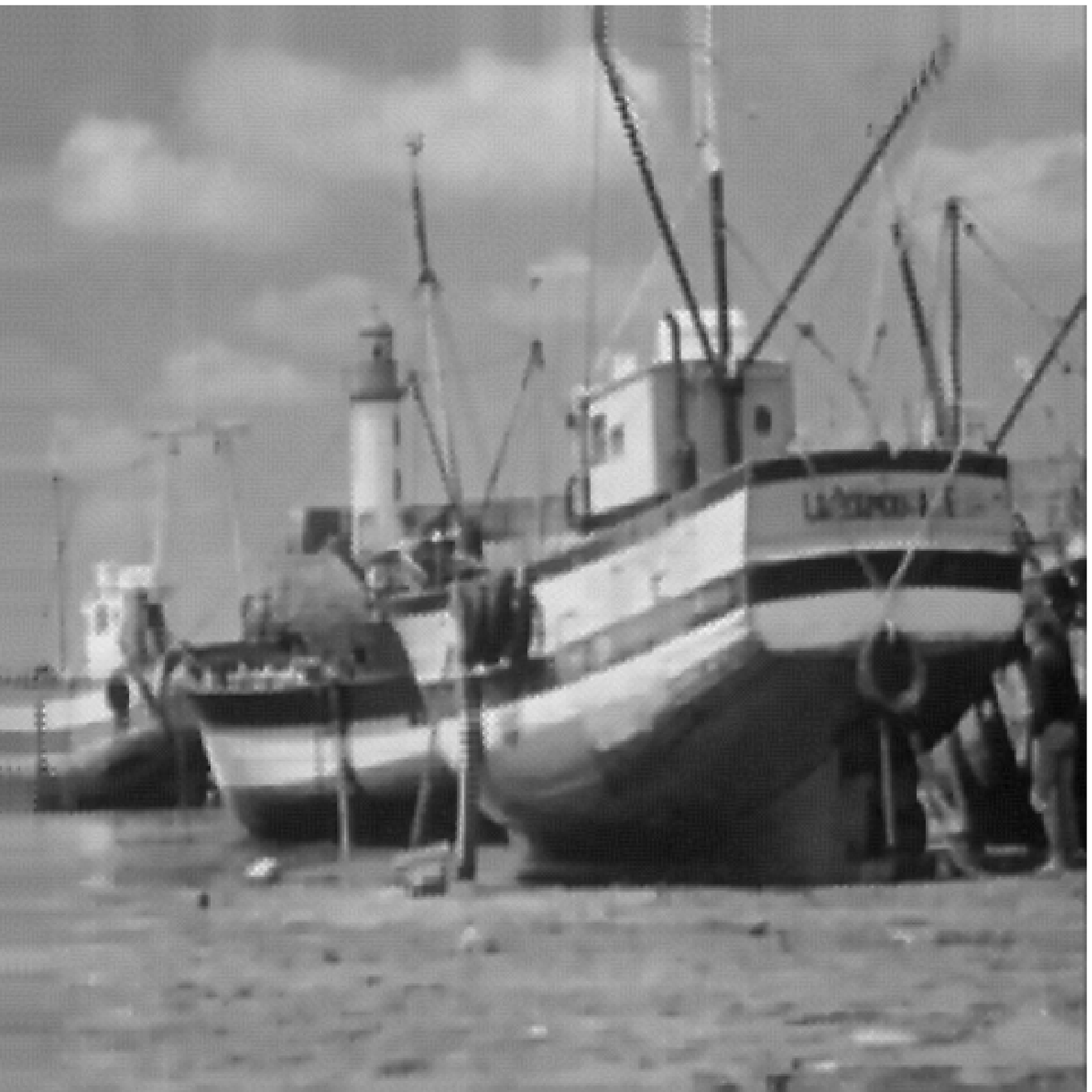}
		\caption{SDR}
		\label{fig:subboat512nScl4-SDR}
	\end{subfigure}
	\begin{subfigure}[t]{0.28\textwidth}
		\centering
		\includegraphics[width=\textwidth,trim={382pt 250pt 10pt 142pt},clip]{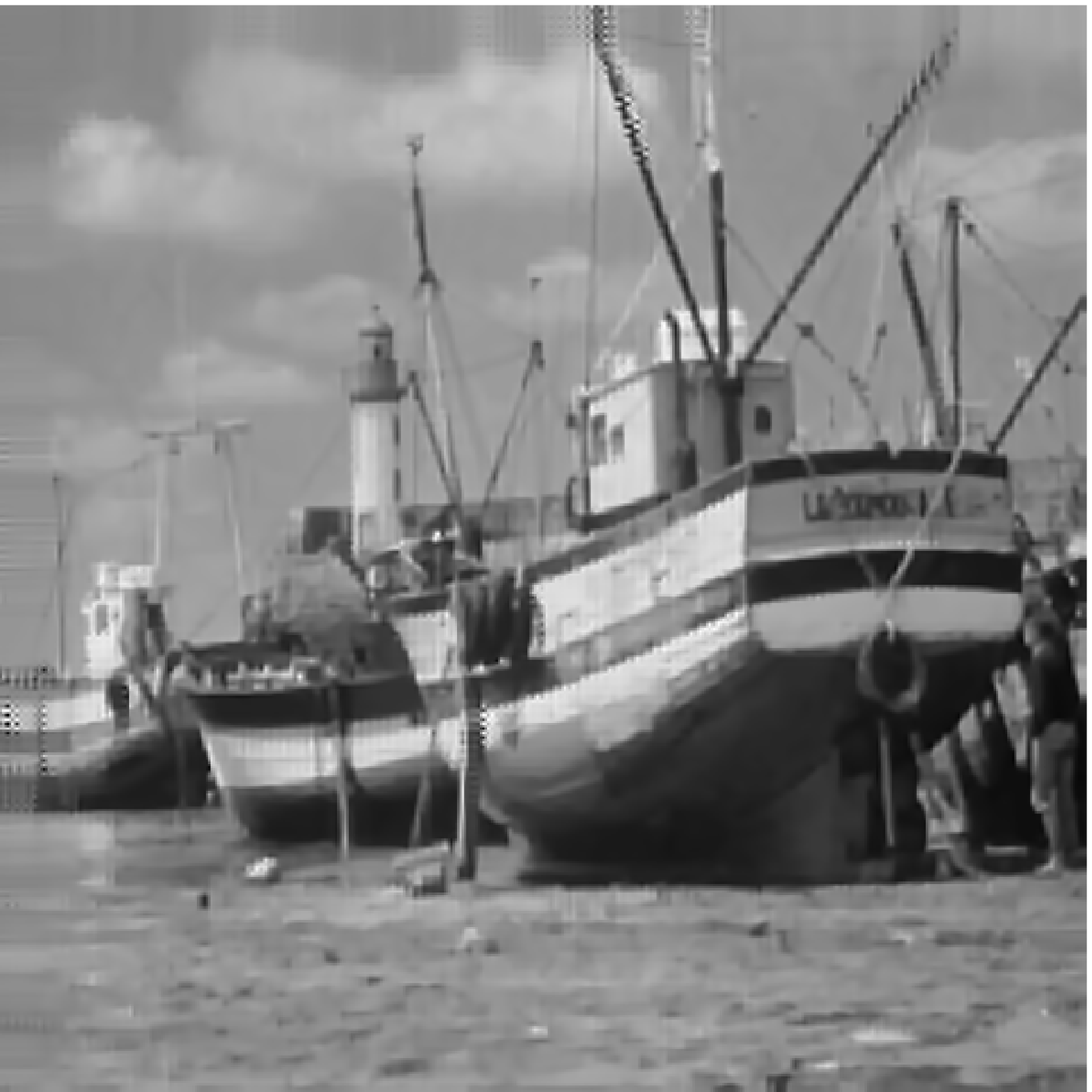}
		\caption{Nuclear}
		\label{fig:subboat512nScl4-NT23}
	\end{subfigure}
	\begin{subfigure}[t]{0.28\textwidth}
		\centering
		\includegraphics[width=\textwidth,trim={382pt 250pt 10pt 142pt},clip]{boat512-HR.eps}
		\caption{True HR image}
		\label{fig:subboat512nScl4-HR}
	\end{subfigure}
	\caption{Zoom-in comparison of different algorithms on ``Boat'' image for $r=4$.
		(a) The reference LR image.
		(b) Result of the TF model \cite{chan2007framelet}.
		(c) Result of the MAP model \cite{ma2015multi}.
		(d) Result of the SDR model \cite{li2010multiframe}.
		(e) Result of our nuclear model ($\lambda=1, \rho=400$).
		(f) Zoomed-in original HR image.}
	\label{fig:subboat512nScl4results}
\end{figure*}

\subsection{Real videos}

In the following,  experiments on real videos are carried out. Four videos ``Text'',  ``Disk'',  ``Alpaca'' and ``Eia''  are downloaded from the website:

\href{https://users.soe.ucsc.edu/\textasciitilde milanfar/software/sr-datasets.html}{https://users.soe.ucsc.edu/\textasciitilde milanfar/software/sr-datasets.html}.

\noindent
The basic information of these videos are listed in Table \ref{tab:basicinfoofvideos}.
We see that they are very low-resolution videos. Fig. \ref{fig:LRimages} shows the reference LR images for these videos. It is difficult to discern most of the letters from the reference
images.

\begin{figure*}[htp]
    \centering
    \begin{subfigure}[t]{0.22\textwidth}
        \centering
        \includegraphics[width=\textwidth]{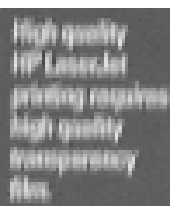}
        \caption{Text}
        \label{fig:text-LR}
    \end{subfigure}
    \begin{subfigure}[t]{0.22\textwidth}
    	\centering
    	\includegraphics[width=\textwidth]{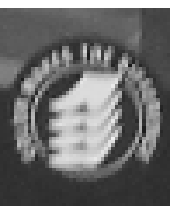}
    	\caption{Disk}
    	\label{fig:disk-LR}
    \end{subfigure}
    \begin{subfigure}[t]{0.22\textwidth}
        \centering
        \includegraphics[width=\textwidth,trim={23pt 0pt 22pt 0pt},clip]{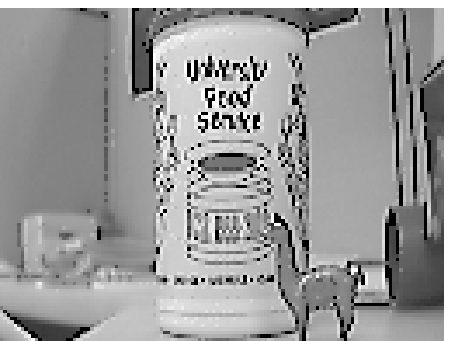}
        \caption{Alpaca}
        \label{fig:Alpaca-LR}
    \end{subfigure}
    \begin{subfigure}[t]{0.22\textwidth}
        \centering
        \includegraphics[width=\textwidth,height=1.16\textwidth,trim={4pt 0pt 2pt 0pt},clip]{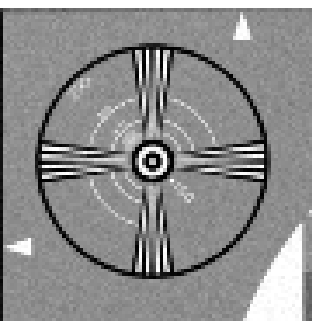}
        \caption{Eia}
        \label{fig:eia-LR}
    \end{subfigure}
\caption{
	The reference  LR images of (a) ``Text'',
	(b)  ``Disk'',
	(c)  ``Alpaca'', and
	(d) ``Eia''.
}\label{fig:LRimages}
\end{figure*}

The first test video is the ``Text Video''. The results are shown in Fig. \ref{fig:textnScl2results}. We see that the TF model produces blurry reconstructions. The images by the MAP model have obvious distortions.  We also see that for the SDR model, some of the letters are coalesced, e.g. the word  ``film''.  The results of  the nuclear model  is better. One can easily tell each word and there are no obvious artifacts for the letters.

\begin{figure*}[htp]
    \centering
    \begin{subfigure}[t]{0.22\textwidth}
    	\centering
    	\includegraphics[width=\textwidth]{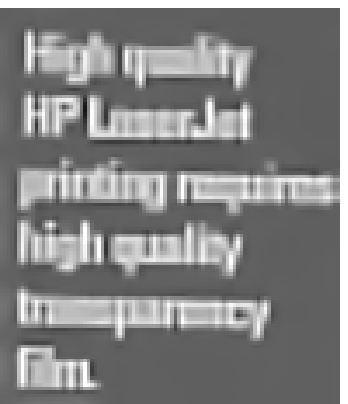}
    \end{subfigure}
    \begin{subfigure}[t]{0.22\textwidth}
        \centering
        \includegraphics[width=\textwidth]{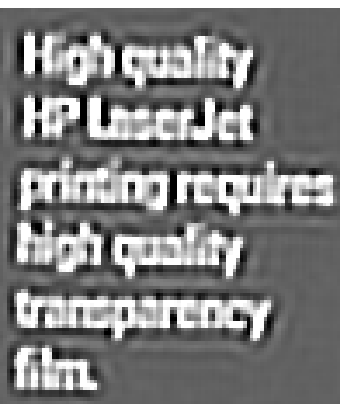}
    \end{subfigure}
    \begin{subfigure}[t]{0.22\textwidth}
        \centering
        \includegraphics[width=\textwidth]{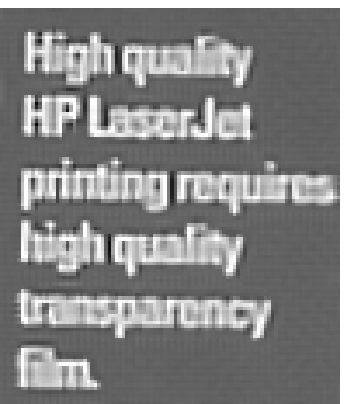}
    \end{subfigure}
    \begin{subfigure}[t]{0.22\textwidth}
        \centering
        \includegraphics[width=\textwidth]{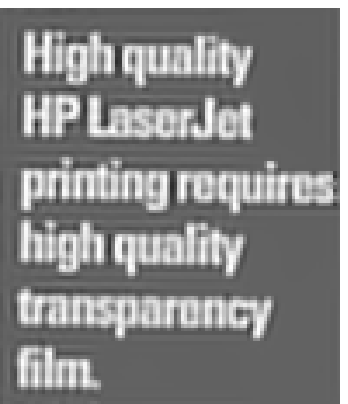}
    \end{subfigure}
    	\begin{subfigure}[t]{0.22\textwidth}
		\centering
		\includegraphics[width=\textwidth]{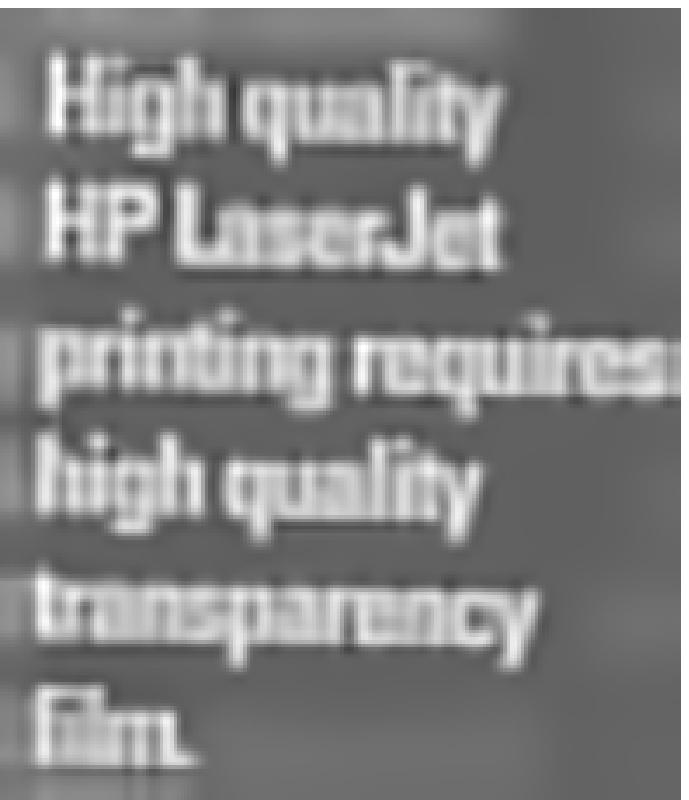}
		\caption{TF}
		\label{fig:textnScl4-TF}
	\end{subfigure}
	\begin{subfigure}[t]{0.22\textwidth}
		\centering
		\includegraphics[width=\textwidth]{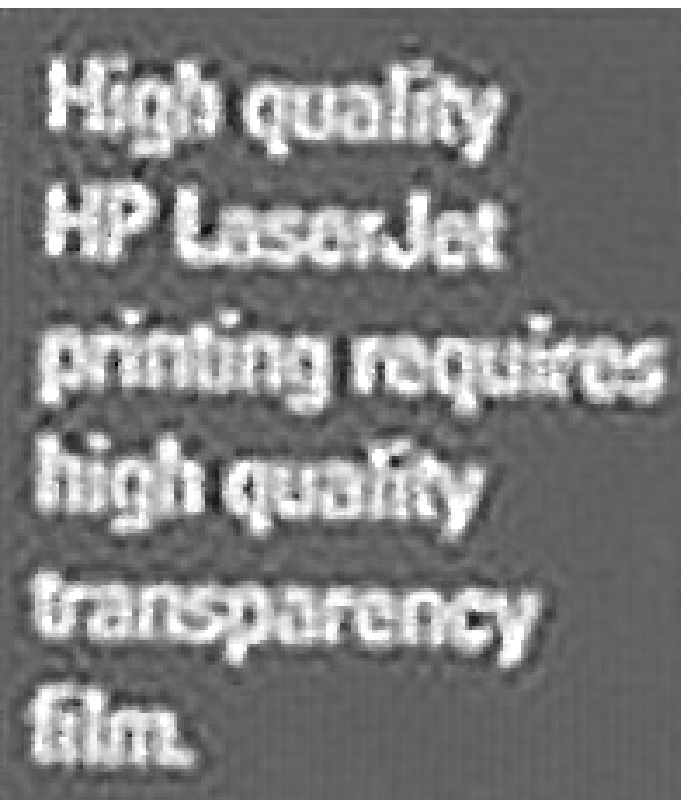}
		\caption{MAP}
		\label{fig:textnScl4-MZ}
	\end{subfigure}
	\begin{subfigure}[t]{0.22\textwidth}
		\centering
		\includegraphics[width=\textwidth]{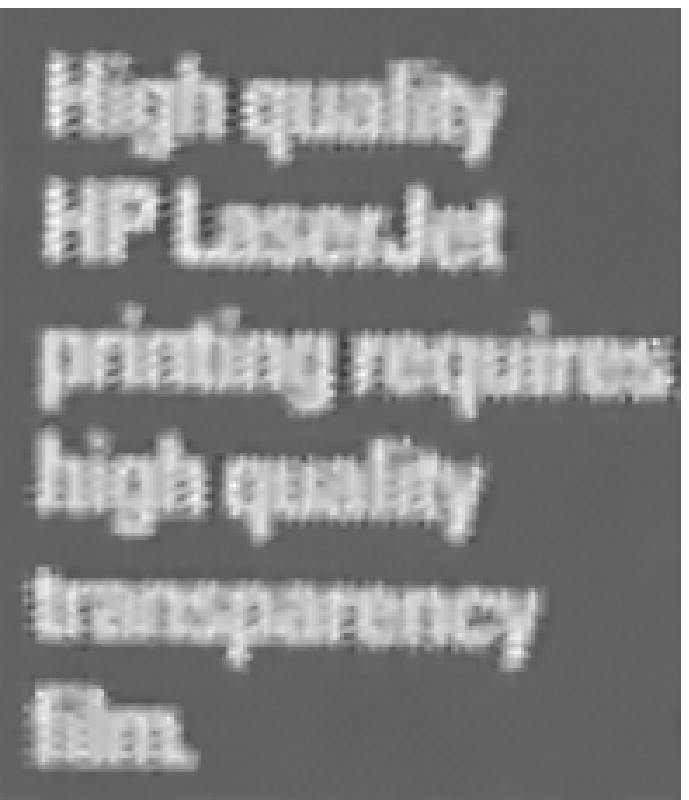}
		\caption{SDR}
		\label{fig:textnScl4-SDR}
	\end{subfigure}
	\begin{subfigure}[t]{0.22\textwidth}
		\centering
		\includegraphics[width=\textwidth]{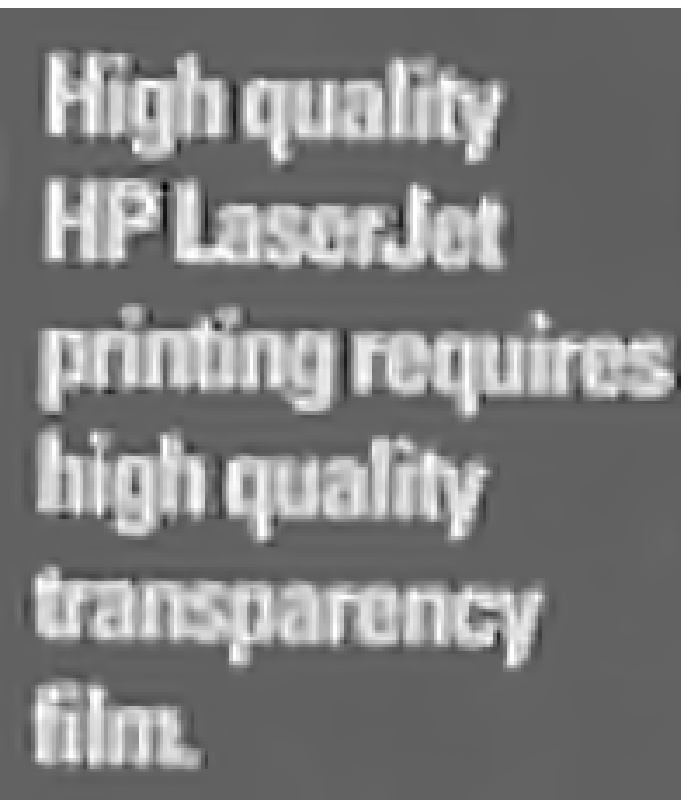}
		\caption{Nuclear }
		\label{fig:textnScl4-NT23}
	\end{subfigure}
\caption{Comparison of different algorithms on ``Text Video''.
Top row with upsampling factor $r=2$ and second row with $r=4$.
	(a) Result of  the TF model \cite{chan2007framelet}.
	(b) Result of  the MAP model \cite{ma2015multi}.
	(d) Result of  the SDR model \cite{li2010multiframe}.
	(d) Result of  our nuclear model ($\lambda=1.5,  \rho=50$ for $r=2$ and
$\lambda=1.375,  \rho=60$ for $r=4$).
}
\label{fig:textnScl2results}
\end{figure*}

The second video is the ``Disk Video'', which contains 26 gray-scale images
with the last 7  ones being zoom-in images. So we only use the first 19 frames in our experiment. The  results are shown in Fig. \ref{fig:disknScl2results}. The TF model again produces blurry reconstructions.
The MAP results are better but still blurry. The SDR results have some artifacts
especially in the word ``DIFFERENCE''. Our results are the
best ones with each letter being well reconstructed, especially when $r=2$.

\begin{figure*}[htp]
  \centering
  \begin{subfigure}[t]{0.22\textwidth}
  	\centering
  	\includegraphics[width=\textwidth]{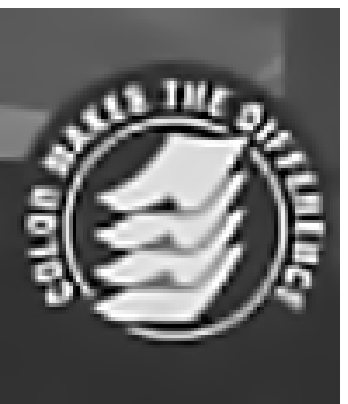}
  \end{subfigure}
  \begin{subfigure}[t]{0.22\textwidth}
    \centering
    \includegraphics[width=\textwidth]{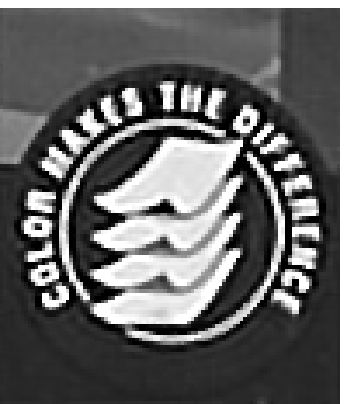}
  \end{subfigure}
  \begin{subfigure}[t]{0.22\textwidth}
    \centering
    \includegraphics[width=\textwidth]{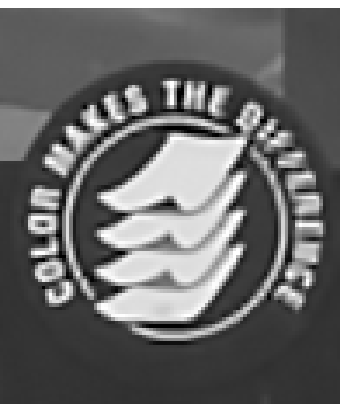}
  \end{subfigure}
  \begin{subfigure}[t]{0.22\textwidth}
    \centering
    \includegraphics[width=\textwidth]{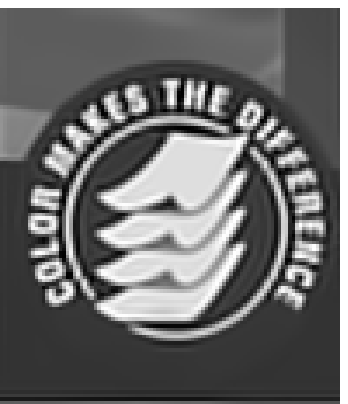}
  \end{subfigure}
  	\begin{subfigure}[t]{0.22\textwidth}
		\centering
		\includegraphics[width=\textwidth]{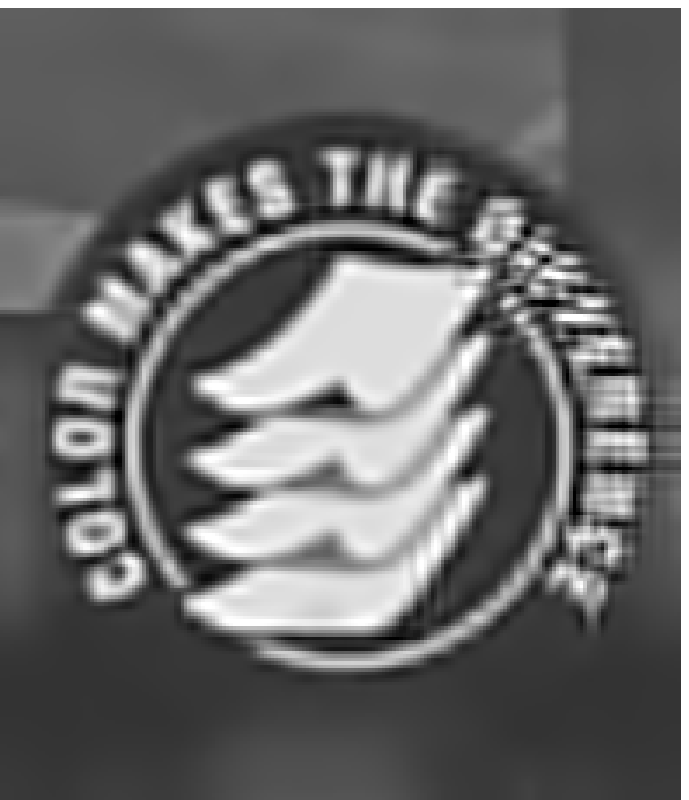}
		\caption{TF}
		\label{fig:disknScl4-TF}
	\end{subfigure}
	\begin{subfigure}[t]{0.22\textwidth}
		\centering
		\includegraphics[width=\textwidth]{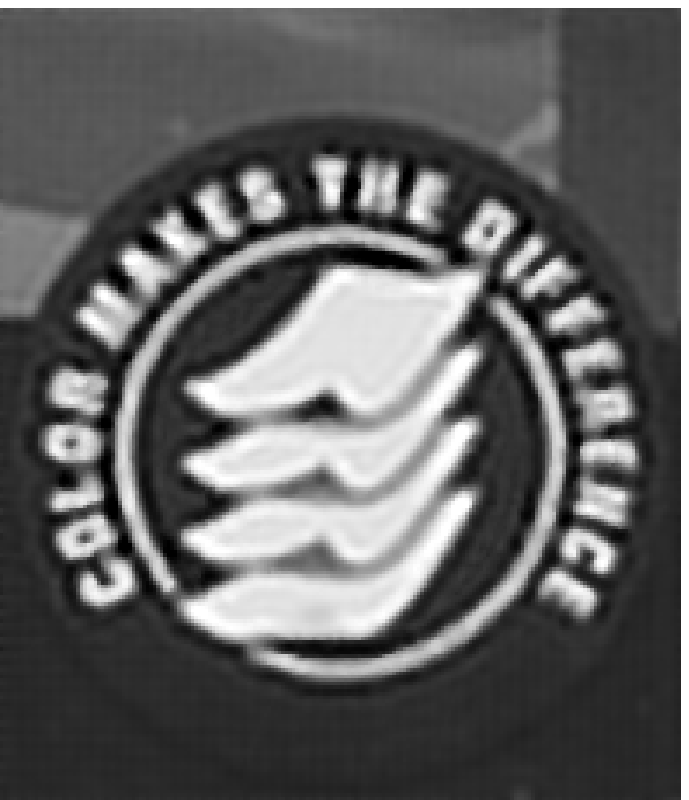}
		\caption{MAP}
		\label{fig:disknScl4-MZ}
	\end{subfigure}
	\begin{subfigure}[t]{0.22\textwidth}
		\centering
		\includegraphics[width=\textwidth]{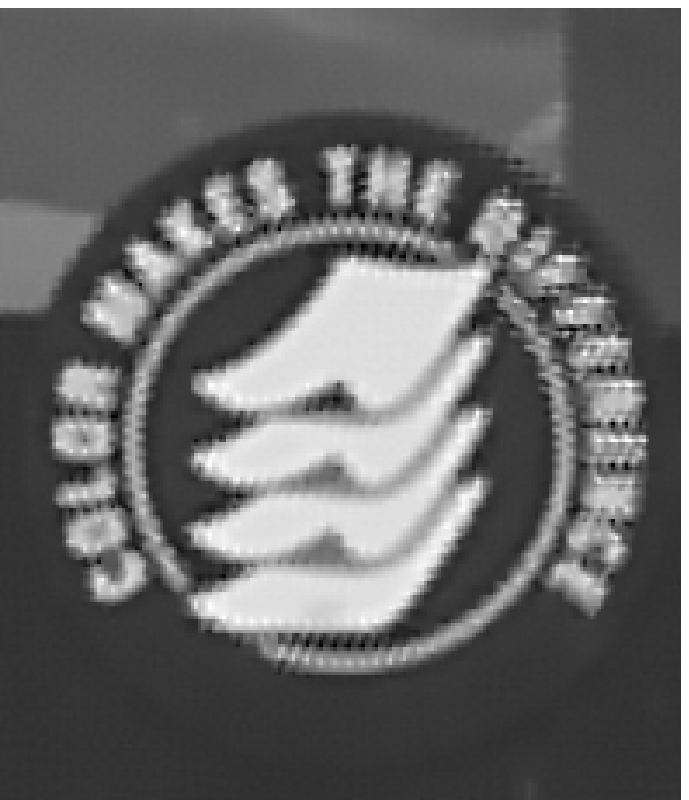}
		\caption{SDR}
		\label{fig:disknScl4-SDR}
	\end{subfigure}
	\begin{subfigure}[t]{0.22\textwidth}
		\centering
		\includegraphics[width=\textwidth]{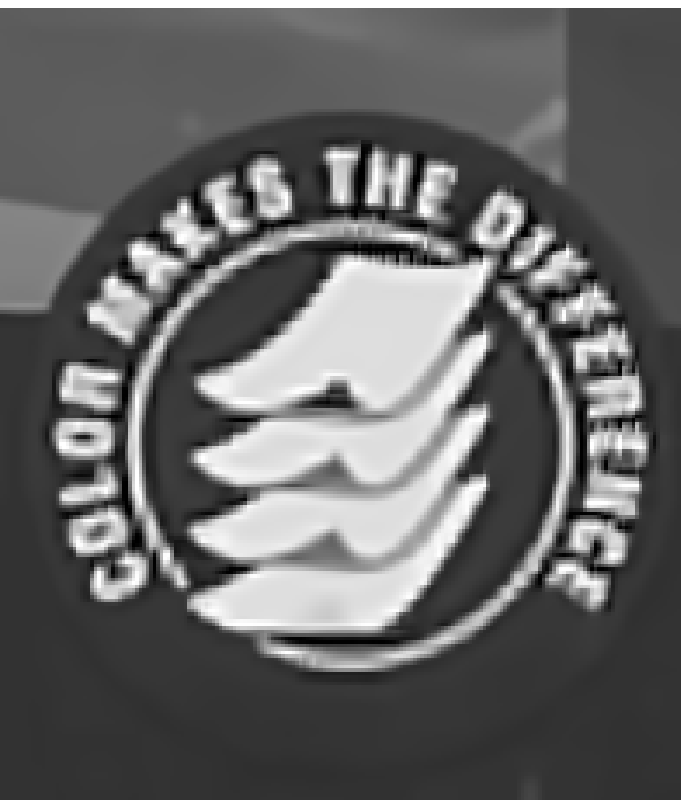}
		\caption{Nuclear}
		\label{fig:disknScl4-NT23}
	\end{subfigure}
  \caption{Comparison of different algorithms on ``Disk Video''.
  Top row with upsampling factor $r=2$ and second row with $r=4$.
  	(a) Result of the TF model \cite{chan2007framelet}.
  	(b) Result of the MAP model \cite{ma2015multi}.
  	(c) Result of the SDR model \cite{li2010multiframe}.
  	(d) Result of our nuclear model ($\lambda=1.125, \rho=50$ for both $r=2$ and  $4$).
  }
  \label{fig:disknScl2results}
\end{figure*}

The third video is the ``Alpaca Video'', and the results are shown in Fig. \ref{fig:AlpacanScl2results}.
When $r=2$, the word ``Service'' are not clear from the TF model, the MAP model and the SDR model. When $r=4$, the resulting images from all models are improved and
the phrase ``University Food Service'' is clearer. However we can see that our nuclear model still gives the best reconstruction.

\begin{figure*}[htp!]
  \centering
	\begin{subfigure}[t]{0.24\textwidth}
		\centering
		\includegraphics[width=\textwidth,trim={46pt 0pt 44pt 0pt},clip]{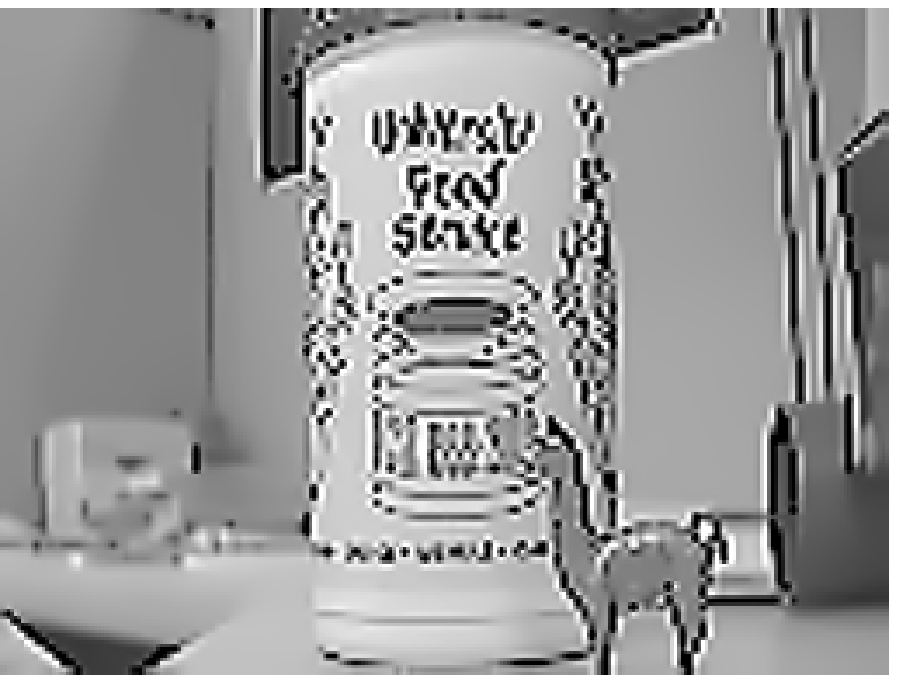}
	\end{subfigure}
  \begin{subfigure}[t]{0.24\textwidth}
    \centering
    \includegraphics[width=\textwidth,trim={69pt 0pt 67pt 0pt},clip]{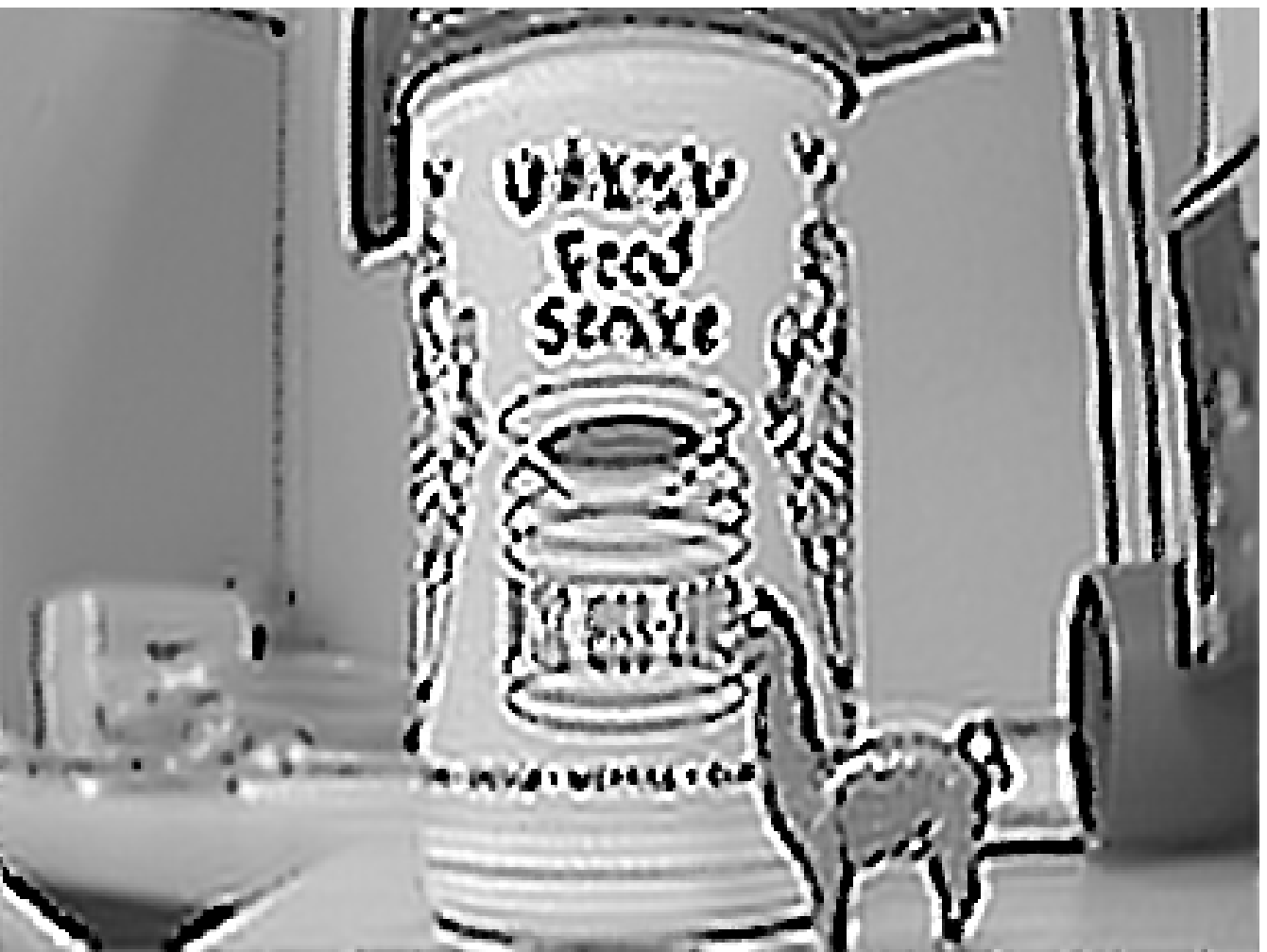}
  \end{subfigure}
  \begin{subfigure}[t]{0.24\textwidth}
    \centering
    \includegraphics[width=\textwidth,trim={46pt 0pt 44pt 0pt},clip]{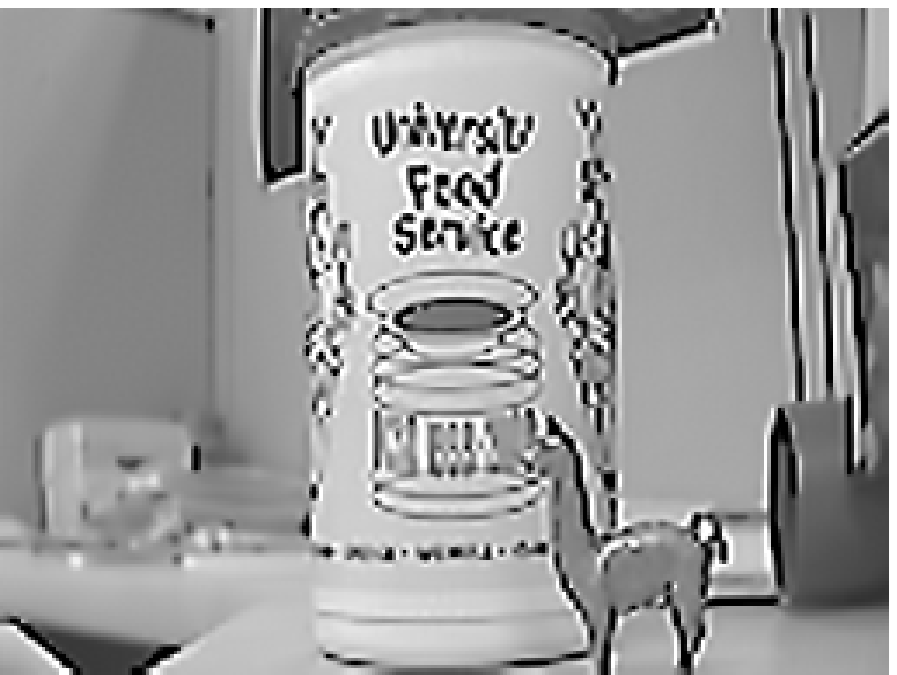}
  \end{subfigure}
  \begin{subfigure}[t]{0.24\textwidth}
    \centering
    \includegraphics[width=\textwidth,trim={46pt 0pt 44pt 0pt},clip]{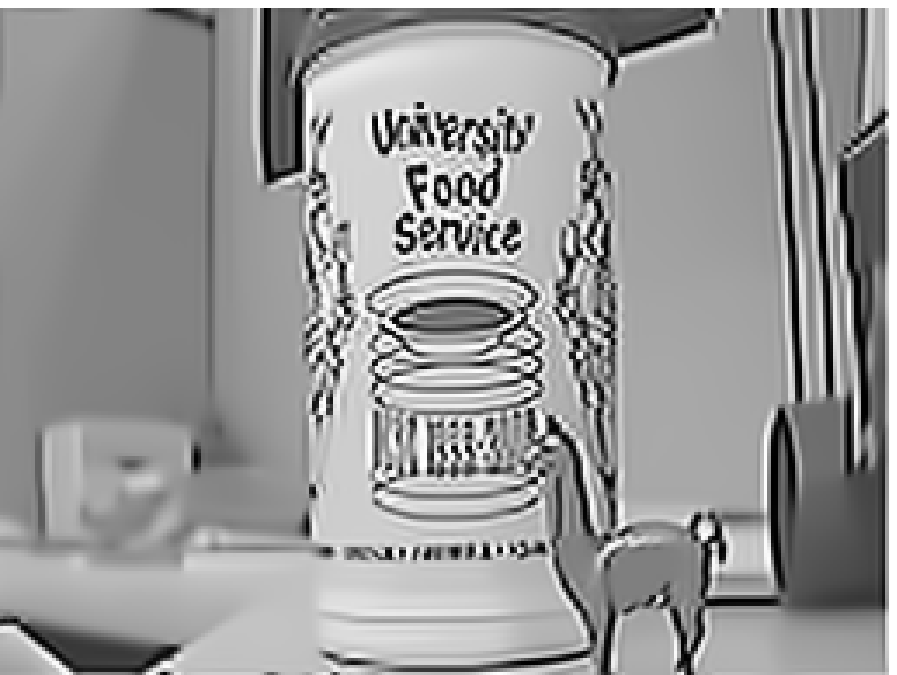}
  \end{subfigure}
  	\begin{subfigure}[t]{0.24\textwidth}
		\centering
		\includegraphics[width=\textwidth,trim={92pt 0pt 90pt 0pt},clip]{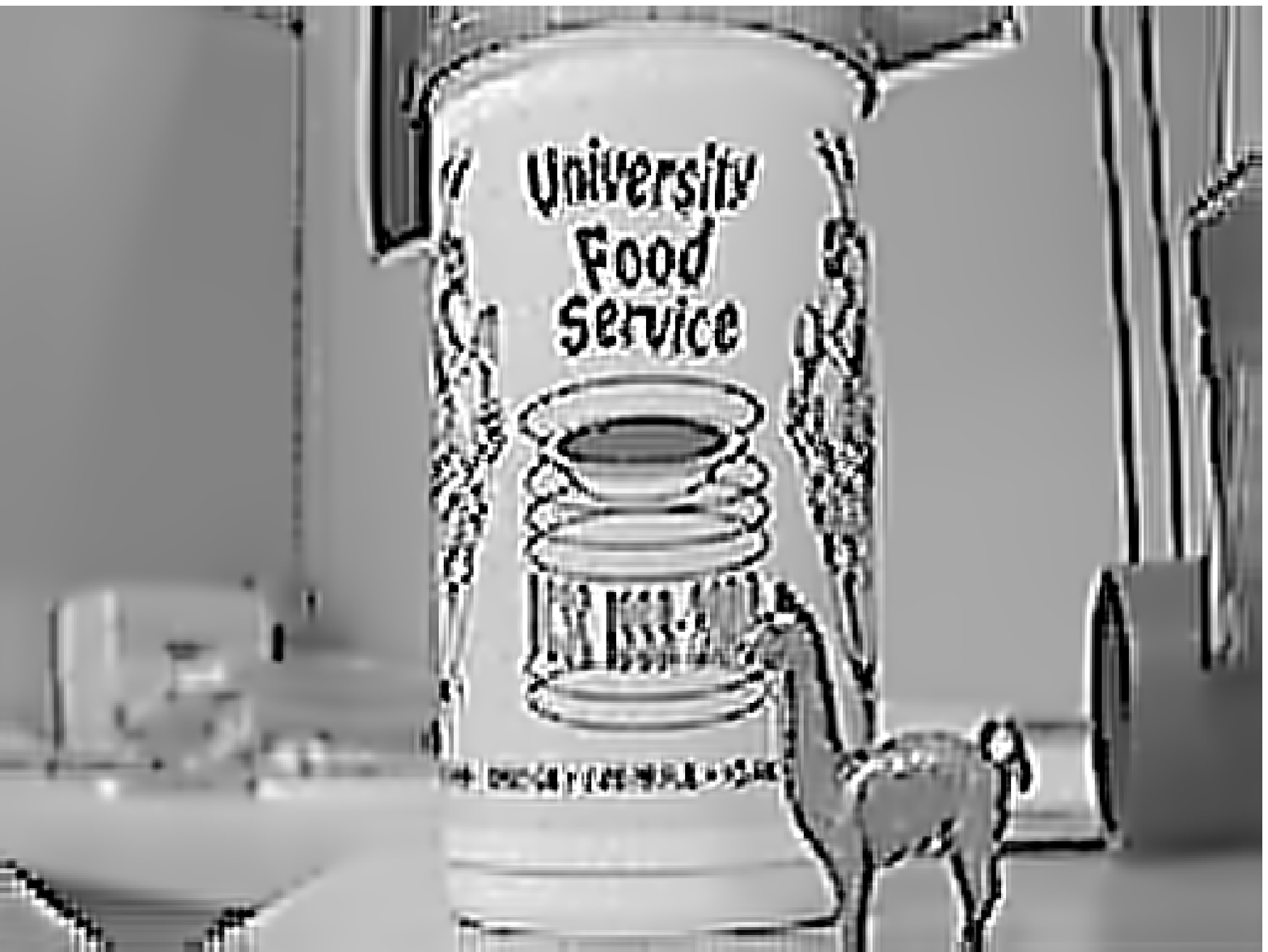}
		\caption{TF}
		\label{fig:AlpacanScl4-TF}
	\end{subfigure}
	\begin{subfigure}[t]{0.24\textwidth}
		\centering
		\includegraphics[width=\textwidth,trim={92pt 0pt 90pt 0pt},clip]{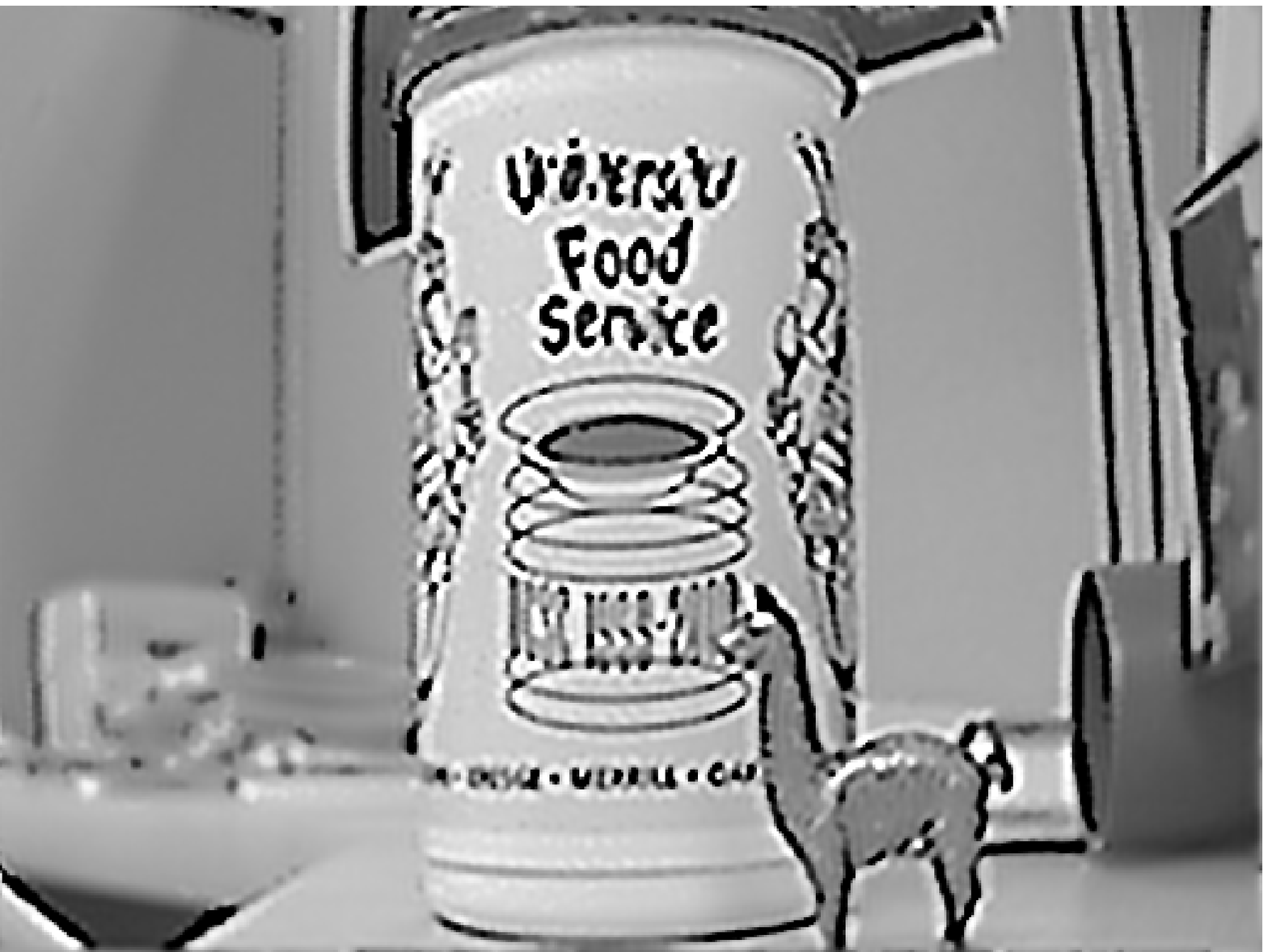}
		\caption{MAP}
		\label{fig:AlpacanScl4-MZ}
	\end{subfigure}
	\begin{subfigure}[t]{0.24\textwidth}
		\centering
		\includegraphics[width=\textwidth,trim={92pt 0pt 90pt 0pt},clip]{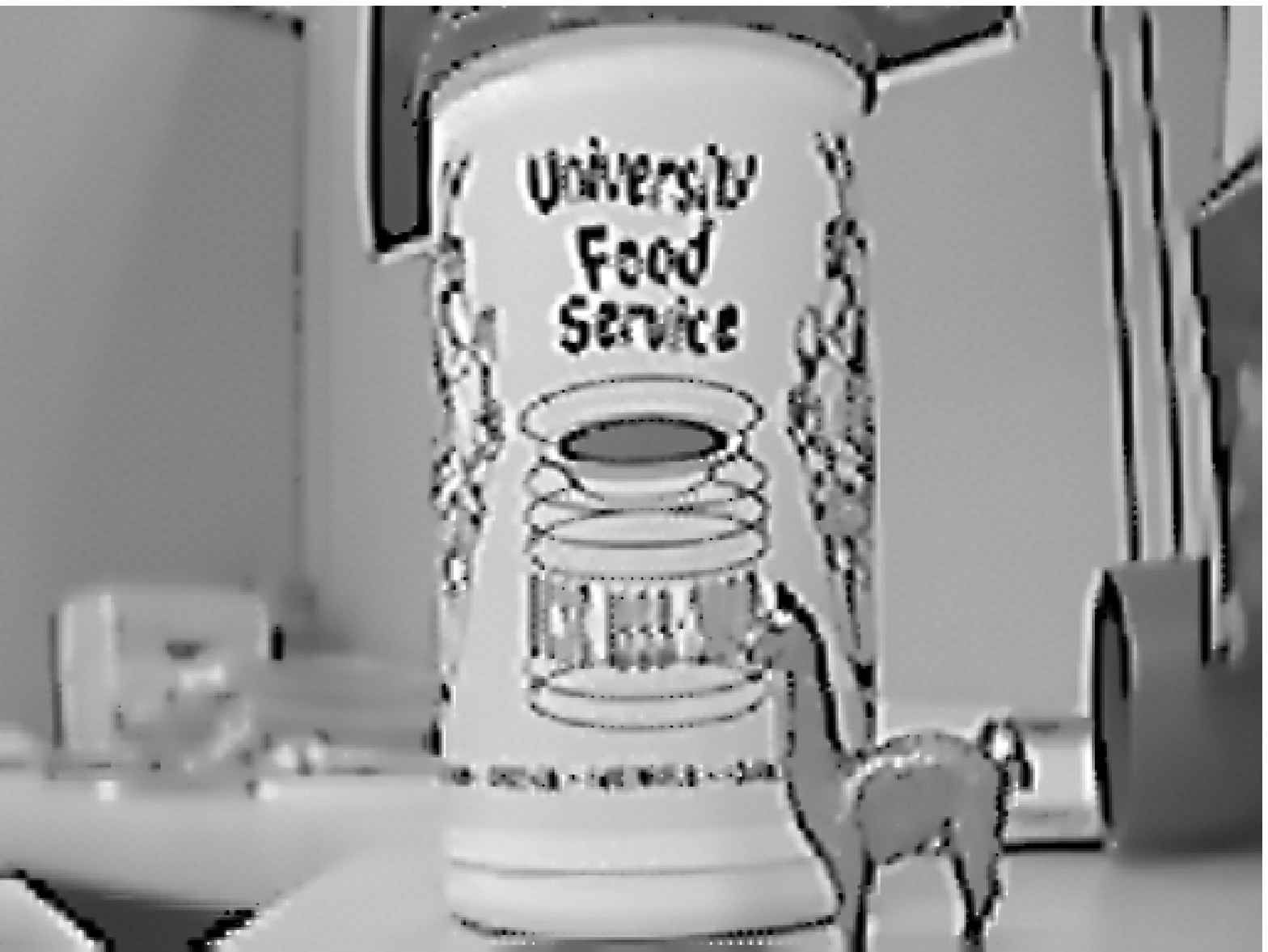}
		\caption{SDR}
		\label{fig:AlpacanScl4-SDR}
	\end{subfigure}
	\begin{subfigure}[t]{0.24\textwidth}
		\centering
		\includegraphics[width=\textwidth,trim={92pt 0pt 90pt 0pt},clip]{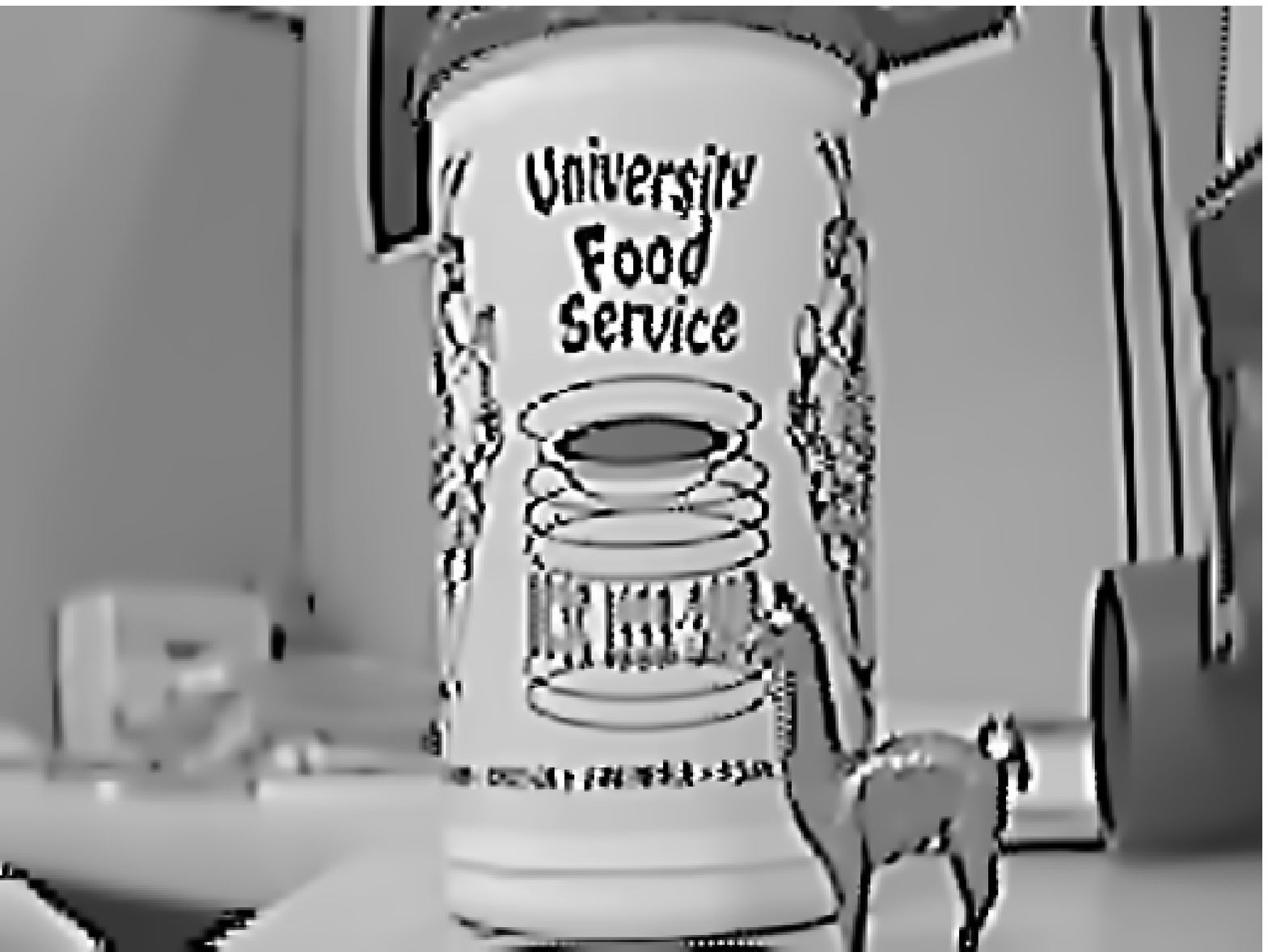}
		\caption{Nuclear }
		\label{fig:AlpacanScl4-NT23}
	\end{subfigure}
   \caption{Comparison of different algorithms on ``Alpaca Video''.
  Top row with upsampling factor $r=2$ and second row with $r=4$.
  	(a) Result of the TF model \cite{chan2007framelet}.
  	(b) Result of the MAP model \cite{ma2015multi}.
  	(c) Result of the SDR model \cite{li2010multiframe}.
  	(d) Result of our nuclear model ($\lambda=1, \rho=50$ for $r=2$
  and $\lambda=0.8, \rho=50$ for $r=4$).
   }
  \label{fig:AlpacanScl2results}
\end{figure*}

The fourth video is the ``Eia Video'' which show a testing image.
There are some concentric circles labeled with different numbers
in decreasing sizes. The results for $r=4$ are shown in  Fig. \ref{fig:eianScl4results}.
Our method gives an image where one can discern the number up to "500" with
almost no artifacts while all the other methods can discern up to "200" at best
with some noise or edge artifacts. This example clearly demonstrates
the effectiveness of our model in MFSR.

\begin{figure*}[htp!]
	\centering
	\begin{subfigure}[t]{0.24\textwidth}
		\centering
		\includegraphics[width=\textwidth]{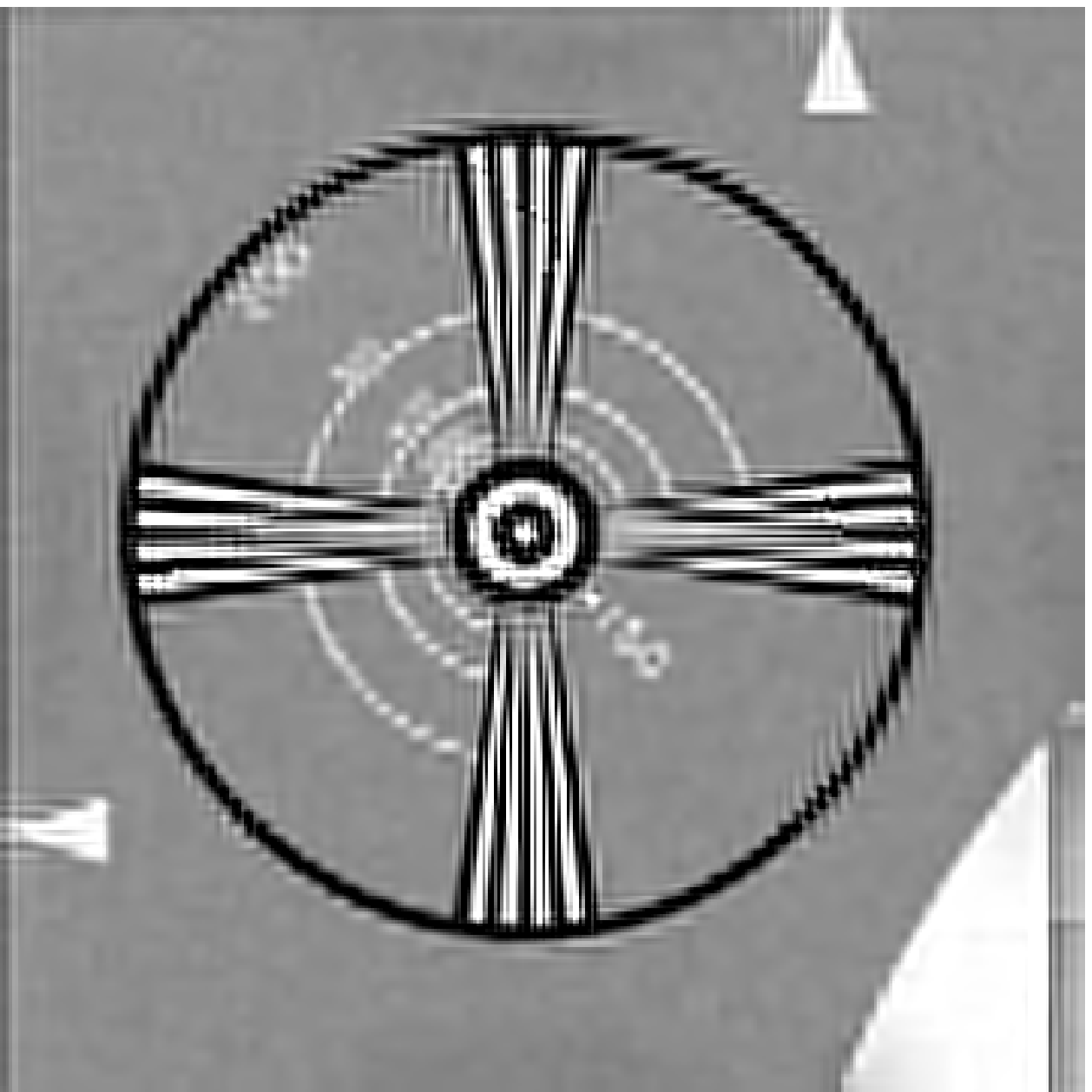}
		\caption{TF}
		\label{fig:eianScl4-TF}
	\end{subfigure}
	\begin{subfigure}[t]{0.24\textwidth}
		\centering
		\includegraphics[width=\textwidth]{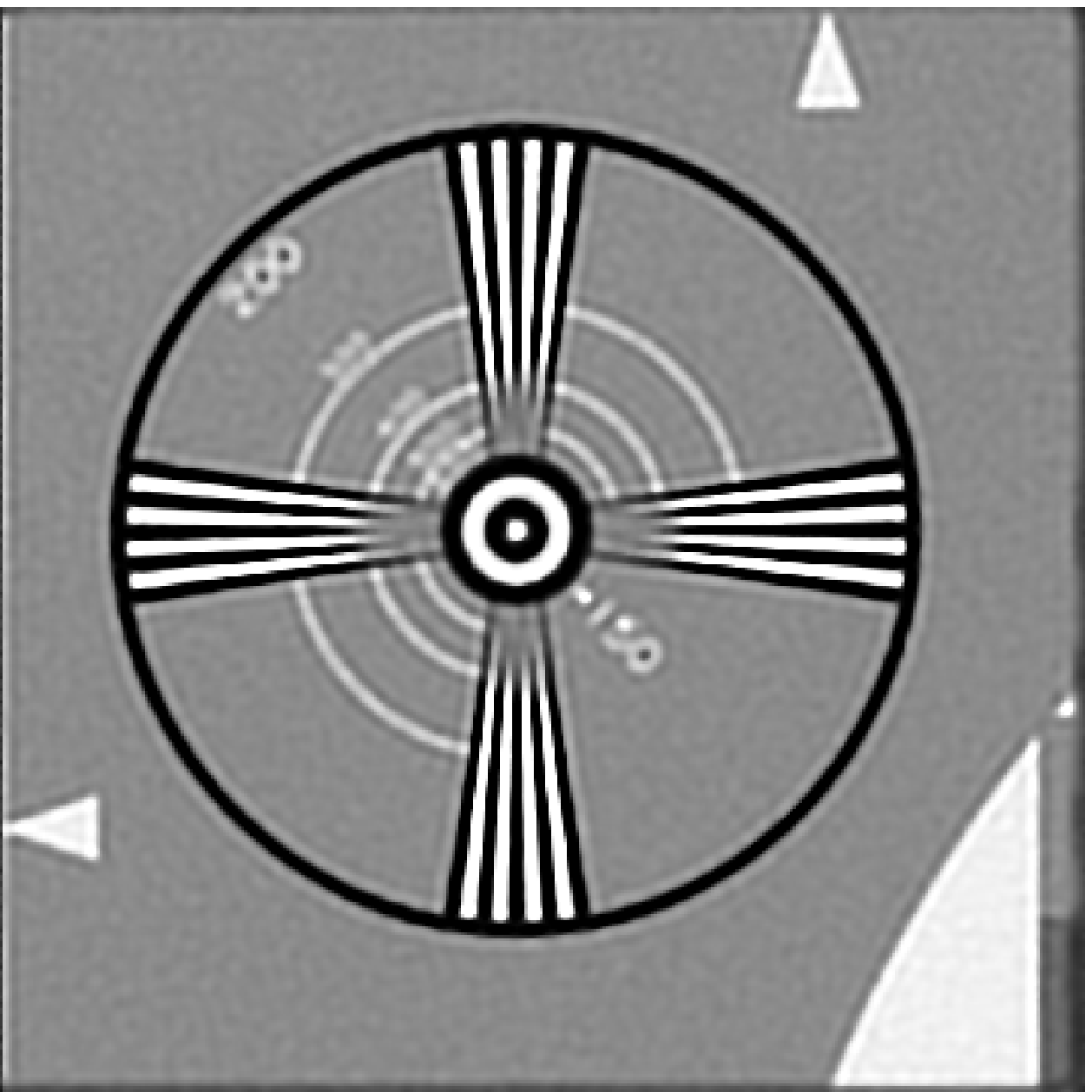}
		\caption{MAP}
		\label{fig:eianScl4-MZ}
	\end{subfigure}
	\begin{subfigure}[t]{0.24\textwidth}
		\centering
		\includegraphics[width=\textwidth]{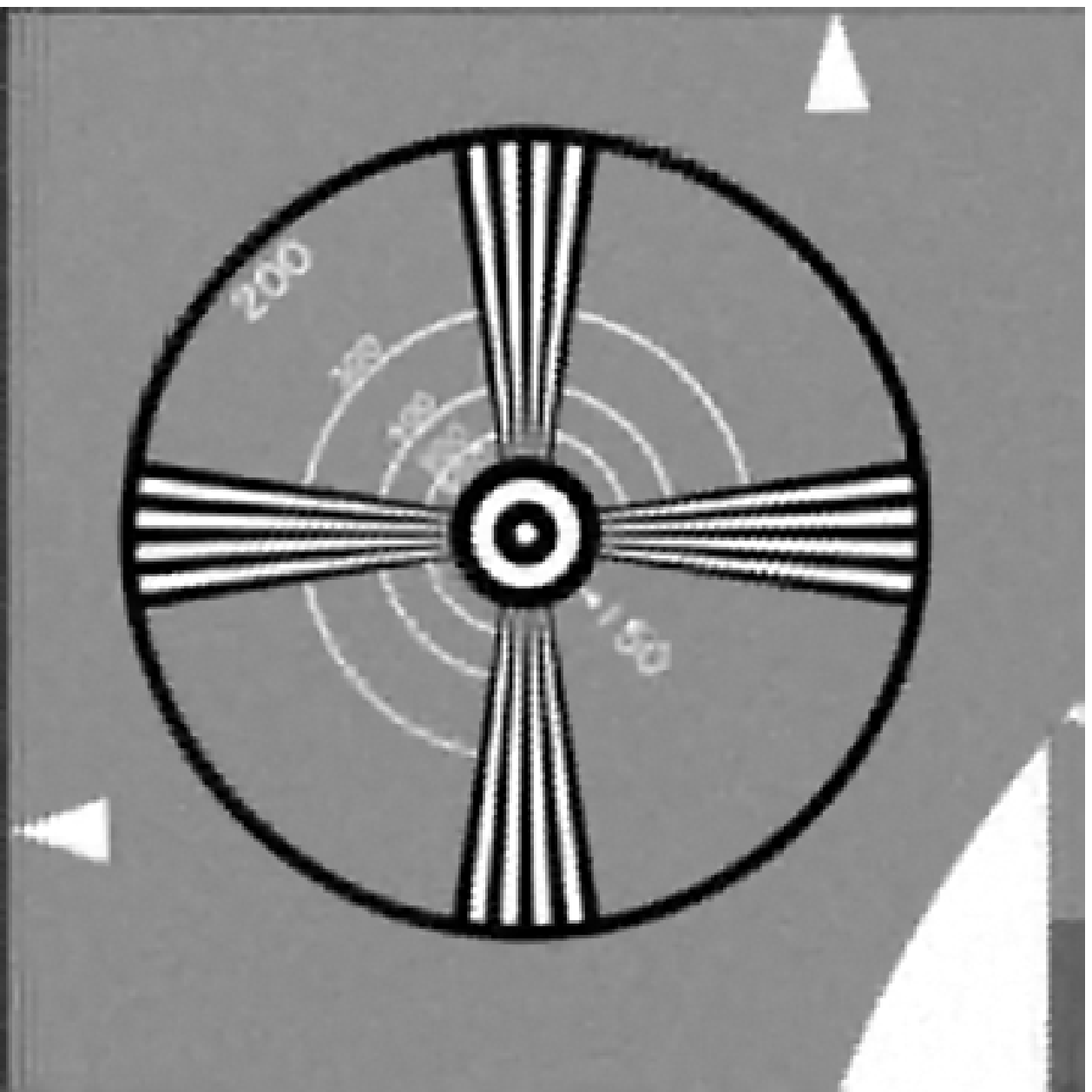}
		\caption{SDR}
		\label{fig:eianScl4-SDR}
	\end{subfigure}
	\begin{subfigure}[t]{0.24\textwidth}
		\centering
		\includegraphics[width=\textwidth]{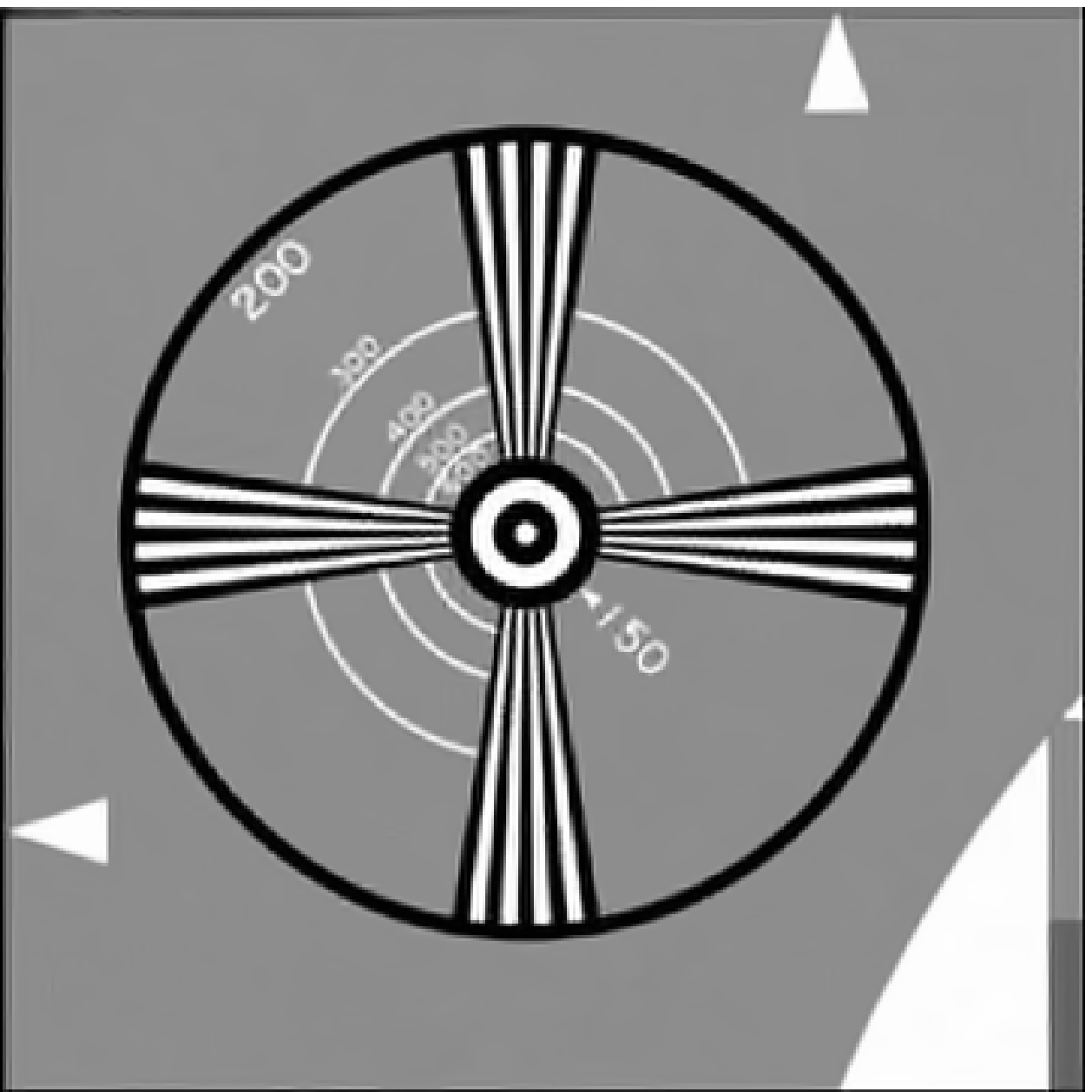}
		\caption{Nuclear }
		\label{fig:eianScl4-NT23}
	\end{subfigure}
	\caption{Comparison of different algorithms on ``Eia Video'' with upsampling factor $r=4$.
		(a) Result of  the TF model \cite{chan2007framelet}.
		(b) Result of  the MAP model \cite{ma2015multi}.
		(c) Result of  the SDR model \cite{li2010multiframe}.
		(d) Result of  our nuclear model ($\lambda=1,  \rho=50$).
	}
	\label{fig:eianScl4results}
\end{figure*}

The last video is a color video which is used in the tests in   \cite{chan2004tight,chan2007framelet}. It contains $257$ color frames. We take the $100$-th frame to be the reference frame, see the leftmost figure in Fig. \ref{fig:subbooks8results}.  Frame $90$ to frame $110$ in the video are used as LR images to enhance the reference image.  We transform the RGB images into the Ycbcr color space, and then apply the algorithms to each color channel. Then we transform the resulting HR images back to the RGB color space.
Figs. \ref{fig:subbooks8results} and  \ref{fig:subbooks5results}
show the zoom-in patches of the resulting images by different models.
In Fig. \ref{fig:subbooks8results}, the patch shows a number ``98'' on the spine of a book.  We see that the TF model gives a reasonable result when compared with MAP and SDR. However, our nuclear model gives the clearest ``98'' with very clean background.
Fig. \ref{fig:subbooks5results} shows the spines of
two other books: ``Fourier Transforms'' and ``Digital Image Processing''.
Again, we see that our nuclear model gives the best reconstruction of the words
with much less noisy artifacts.

\begin{figure*}
    \centering
    \begin{minipage}[t]{0.30\textwidth}
		\centering
        \includegraphics[width=\textwidth]{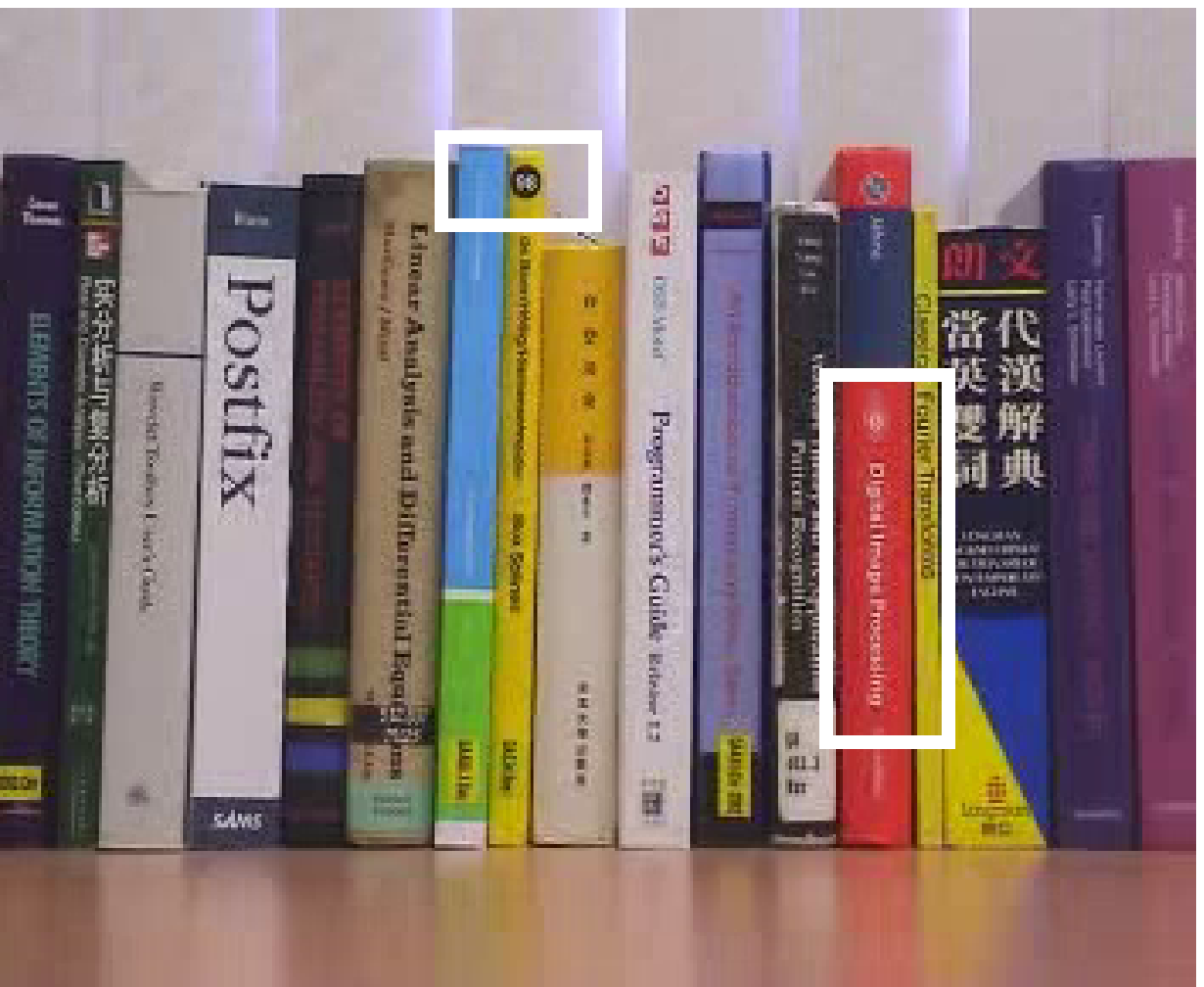}
    \end{minipage}\hfill%
    \begin{minipage}[b]{0.68\textwidth}
	\centering
	\begin{subfigure}[t]{0.32\textwidth}
		\centering
        \includegraphics[width=\textwidth]{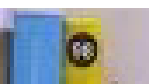}
		\caption{Zoomed-in LR}
		\label{fig:subbooks8-LR}
	\end{subfigure}%
    \hspace{1em}
	\begin{subfigure}[t]{0.31\textwidth}
		\centering
        \includegraphics[width=\textwidth]{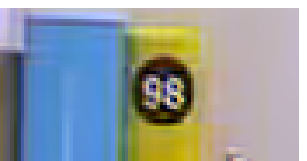}
		\caption{ TF}
		\label{fig:subbooks8-TF}
	\end{subfigure}
	\begin{subfigure}[t]{0.31\textwidth}
	\centering
    \includegraphics[width=\textwidth]{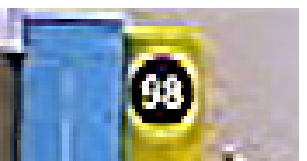}
	\caption{ MAP}
	\label{fig:subbooks8-MZ}
\end{subfigure}
	\begin{subfigure}[t]{0.31\textwidth}
		\centering
        \includegraphics[width=\textwidth]{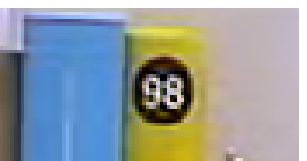}
		\caption{SDR}
		\label{fig:subbooks8-SDR}
	\end{subfigure}
	\begin{subfigure}[t]{0.31\textwidth}
		\centering
		\includegraphics[width=\textwidth]{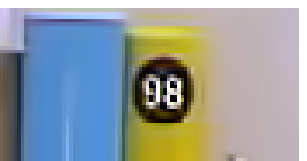}
		\caption{Nuclear}
		\label{fig:subbooks8-NT23}
	\end{subfigure}
   \end{minipage}
	\caption{Zoom-in comparison of different algorithms on ``Books Video'' with $r=2$.
Left-most figure: the LR reference frame with zoom-in areas marked.
		(a) Zoomed-in LR image.
		(b) Result of the TF model \cite{chan2007framelet}.
		(c) Result of the MAP model \cite{ma2015multi}.
		(d) Result of the SDR model \cite{li2010multiframe}.
		(e) Result of our nuclear model ($\lambda=1.375,  \rho=400$).
        }
        \label{fig:subbooks8results}
\end{figure*}

\begin{figure*}
	\centering
	\begin{subfigure}[t]{0.32\textwidth}
		\centering
        \includegraphics[width=\textwidth]{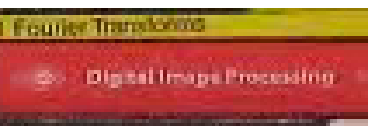}
		\caption{Zoomed-in LR}
		\label{fig:subbooks5-LR}
	\end{subfigure}%
    \hspace{1em}
	\begin{subfigure}[t]{0.31\textwidth}
		\centering
        \includegraphics[width=\textwidth]{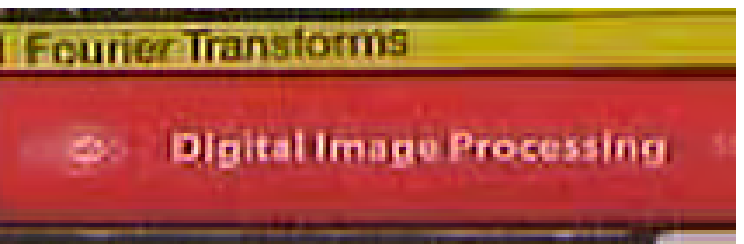}
		\caption{ TF}
		\label{fig:subbooks5-TF}
	\end{subfigure}
	\begin{subfigure}[t]{0.31\textwidth}
	\centering
        \includegraphics[width=\textwidth]{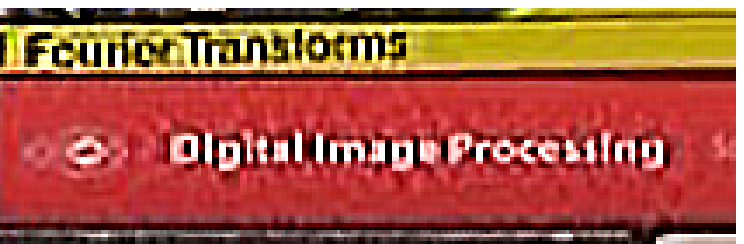}
	\caption{ MAP}
	\label{fig:subbooks5-MZ}
\end{subfigure}
	\begin{subfigure}[t]{0.31\textwidth}
		\centering
        \includegraphics[width=\textwidth]{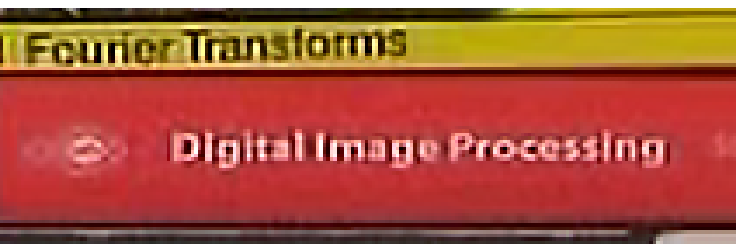}
		\caption{SDR}
		\label{fig:subbooks5-SDR}
	\end{subfigure}
	\begin{subfigure}[t]{0.31\textwidth}
		\centering
        \includegraphics[width=\textwidth]{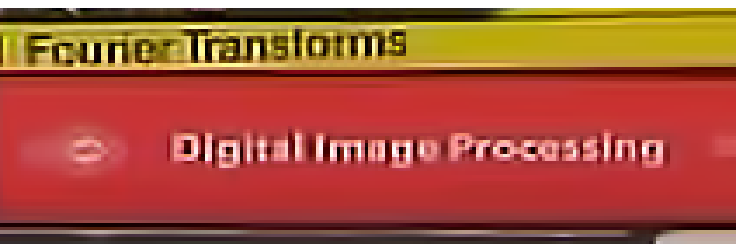}
		\caption{Nuclear}
		\label{fig:subbooks5-NT23}
	\end{subfigure}
	\caption{Another zoom-in comparison on ``Books Video''with $r=2$.
		(a) Zoomed-in LR image.
		(b) Result of  the TF model \cite{chan2007framelet}.
		(c) Result of  the MAP model \cite{ma2015multi}.
		(d) Result of the  SDR model \cite{li2010multiframe}.
		(e) Result of our nuclear model ($\lambda=1.375,  \rho=400$).
        }
        \label{fig:subbooks5results}
\end{figure*}

\section{Conclusion}
\label{sec:conclusion}

In this paper,  we proposed an effective algorithm to reconstruct a high-resolution image using multiple low-resolution images from video clips.  The LR images are first registered to the reference frame by using an optical flow. Then a low-rank model is used to
reconstruct the high-resolution image by making use of the overlapping information between different LR images. Our model can handle complex motions and  illumination changes. Tests on synthetic and real videos show that our model can reconstruct
an HR image with much more details and less artifacts.


\begin{IEEEbiography}{Rui Zhao} received the B.S. degree  in applied mathematics from Tsinghua University, Beijing, China in 2008, the M.S. degree in applied mathematics from Tsinghua University, Beijing, China in 2011. He is now a PhD student of Chinese University of Hong Kong. His main research area is image processing.

\end{IEEEbiography}

\begin{IEEEbiography}{Raymond H. Chan} is a SIAM Fellow and SIAM Council Member. He is
 Choh-Ming Li Professor of Chinese University of Hong Kong and Chairman
of Department of Mathematicsis, Chinese University of Hong Kong.
His research interests include numerical linear
algebra and image processing problems.
\end{IEEEbiography}
%
\end{document}